\RequirePackage{fix-cm}
\documentclass[twocolumn]{svjour3}           
\smartqed  
\usepackage{graphicx}
\usepackage{subcaption}
\usepackage{csquotes}
\usepackage{natbib}
\usepackage{amsmath}
\usepackage{url}
\usepackage{amssymb}

\usepackage{placeins}

\usepackage{booktabs}
\usepackage[table]{xcolor}

\usepackage{array}
\usepackage{makecell}

\begin{document}

\title{Learning Extremal Representations with Deep Archetypal Analysis
}


\author{Sebastian Mathias Keller$^{1}$ \and
        Maxim Samarin$^{1}$            \and
        Fabricio Arend Torres$^{1}$    \and
        Mario Wieser$^{1}$             \and
        Volker Roth$^{1}$ 
}


\institute{Sebastian Mathias Keller \at
           \email{sebastianmathias.keller@unibas.ch}\\
           \\
              $^{1}$ Department of Mathematics and Computer Science,\\
              University of Basel, Basel, Switzerland\\
}

\date{Received: date / Accepted: date}

\maketitle

\begin{abstract}
Archetypes are typical population representatives in an extremal sense, where typicality is understood as the most extreme manifestation of a trait or feature. In linear feature space, archetypes approximate the data convex hull allowing all data points to be expressed as convex mixtures of archetypes. However, it might not always be possible to identify meaningful archetypes in a given feature space. As features are selected a priori, the resulting representation of the data might only be poorly approximated as a convex mixture. Learning an appropriate feature space and identifying suitable archetypes simultaneously addresses this problem. This paper introduces a generative formulation of the linear archetype model, parameterized by neural networks. By introducing the distance-dependent archetype loss, the linear archetype model can be integrated into the latent space of a variational autoencoder, and an optimal representation with respect to the unknown archetypes can be learned end-to-end. 
The reformulation of linear Archetypal Analysis as a variational autoencoder naturally leads to an extension of the model to a deep variational information bottleneck, allowing the incorporation of arbitrarily complex side information during training. As a consequence, the answer to the question "What is typical in a given data set?" can be guided by this additional information. Furthermore, an alternative prior, based on a modified Dirichlet distribution, is proposed. On a theoretical level, this makes the relation to the original archetypal analysis model more explicit, where observations are modelled as samples from a Dirichlet distribution. The real-world applicability of the proposed method is demonstrated by exploring arche\-types of female facial expressions while using multi-rater based emotion scores of these expressions as side information. A second application illustrates the exploration of the chemical space of small organic molecules. In this experiment, it is demonstrated that exchanging the side information but keeping the same set of molecules, e. g. using as side information the heat capacity of each molecule instead of the band gap energy, will result in the identification of different archetypes.  As an application, these learned representations of chemical space might reveal distinct starting points for de novo molecular design.
\end{abstract}


\section{Introduction}
\label{sec:1}
\begin{figure}
\centering
\begin{subfigure}{0.42\textwidth}
   \includegraphics[width=1\linewidth]{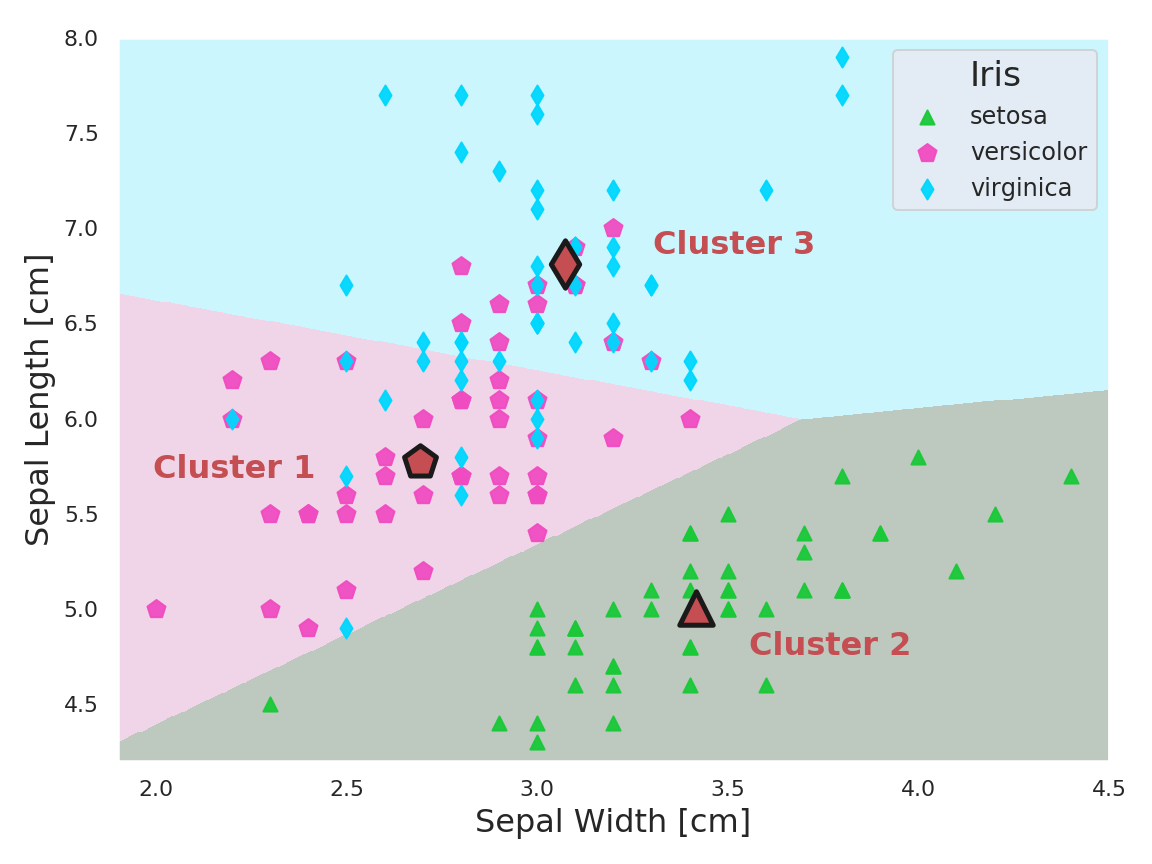}
   \caption{k-Means clustering}
   \label{fig:iris_mn} 
\end{subfigure}
\begin{subfigure}{0.42\textwidth}
   \includegraphics[width=1\linewidth]{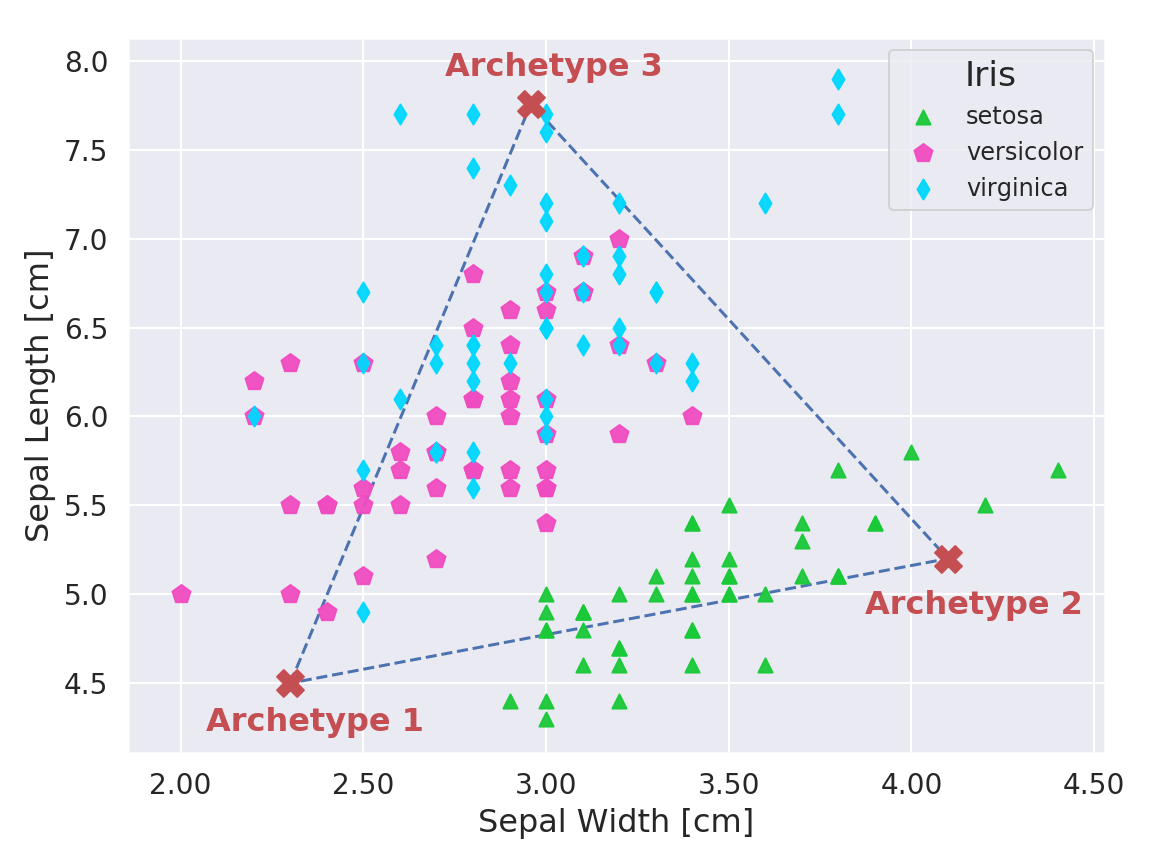}
   \caption{Archetypal Analysis}
   \label{fig:iris_at}
\end{subfigure}
\caption[Prototypes versus Archetypes]{Result of a clustering procedure as well as an archetypal analysis, performed on the Iris data set. For clustering, the k-means algorithm was used, which is an unsupervised clustering algorithm identifying the average representatives of a data set, i. e. the cluster-centroids. Archetypal Analysis on the other hand, seeks to identify extremes in the data set with the goal to represent individual data points as weighted mixtures of these extreme points, the so-called archetypes.}
\label{fig:iris}
\end{figure}

Colloquially, both the words \enquote{archetype} and \enquote{prototype} describe templates or original patterns from which all later forms are developed. However, the concept of a prototype is more common in machine learning and for example encountered as cluster-centroids in classification, where a query point $x$ is assigned to the class of the closest prototype. In an appropriate feature space such a prototype is a typical representative of its class, sharing all traits of the class members, ideally in equal proportion. By contrast, archetypes are characterized as being \textit{extreme points} of the data, such that the complete data set can be well represented as a convex mixture of these extremes or archetypes. Archetypes thus form a polytope approximating the data convex hull. Based on the historic Iris flower data set \citep{anderson1935, fisher1936}, Figure \ref{fig:iris} illustrates the different perspectives both approaches provide in exploring the data. In Figure  \ref{fig:iris_mn} the cluster means as well as the decision boundaries in a 2-dimensional feature space are shown. The clustering was calculated using the k-Means algorithm. Each cluster mean is a typical \textit{average} representative of its respective class, the aforementioned prototype. According to this clustering, the prototypical \textit{Iris virginica} has a sepal width of 3.1cm and a sepal length of 6.8cm. On the other hand, Figure \ref{fig:iris_at} shows the positions of the three archetypal Iris flowers, which are typical \textit{extreme} representatives. The archetypal \textit{Iris virginica} has a sepal width of 3.0cm and a sepal length of 7.8cm. All flowers within the simplex are characterized as weighted mixtures of these archetypes while, in terms of convex mixtures, the optimal location of flowers \textit{outside} the simplex are normal projections onto its surface. In general, a clustering approach is more natural if the existence of a cluster structure can be presumed. Otherwise, archetypal analysis might offer an interesting perspective for exploratory data analysis.
%
%
%
%
%
\section{Exploring Data Sets Through Archetypes}
\label{sec:2}
Archetypal analysis (AA) was first proposed by \cite{cutlerBreiman1994}. It is a linear procedure where archetypes are selected by minimizing the squared error in representing each individual data point as a mixture of archetypes. Identifying the archetypes involves the minimization of a non-linear least squares loss.
%
\subsection{Archetypal Analysis}
\label{sec:linAA}
Linear AA is a form of non-negative matrix factorization where a matrix $X\in \mathbb{R}^{n\times p}$ of $n$ data vectors is approximated as $X\approx ABX = AZ$ with $A\in \mathbb{R}^{n\times k}$, $B\in \mathbb{R}^{k\times n}$, and usually $k < \min\{n,p\}$. The so-called \textit{archetype matrix} $Z\in \mathbb{R}^{k\times p}$ contains the $k$ archetypes $\mathbf{z}_1,..,\mathbf{z}_j,.., \mathbf{z}_k$ with the model being subject to the following constraints:
%
\begin{equation}\label{eq:constraint_A_B}
a_{ij} \geq 0 \text{ }\wedge\text{ } \sum_{j=1}^{k} a_{ij} = 1, \quad 
b_{ji} \geq 0 \text{ }\wedge\text{ } \sum_{i=1}^{n} b_{ji} = 1
\end{equation}
Constraining the entries of $A$ and $B$ to be non-negative and demanding that both weight matrices are row stochastic implies a representation of the data vectors $\mathbf{x}_{i=1..n}$ as a weighted sum of the rows of $Z$ while simultaneously representing the archetypes $\mathbf{z}_{j=1..k}$ themselves as a weighted sum of the $n$ data vectors in $X$:
\begin{equation} \label{eq:at_decomp}
\mathbf{x}_i \approx \sum_{j=1}^{k} a_{ij} \mathbf{z}_j = \mathbf{a}_i Z, \quad
\mathbf{z}_j = \sum_{i=1}^{n} b_{ji} \mathbf{x}_i  = \mathbf{b}_j X
\end{equation}
Due to the constraints on $A$ and $B$ in Eq. \ref{eq:constraint_A_B} both the representation of $\mathbf{x}_i$ and $\mathbf{z}_j$ in Eq. \ref{eq:at_decomp} are \textit{convex} combinations. Therefore the archetypes approximate the data convex hull and increasing the number $k$ of archetypes improves this approximation. The central problem of AA is finding the weight matrices $A$ and $B$ for a given data matrix $X$ and a given number $k$ of archetypes. The non-linear optimization problem consists in minimizing the following residual sum of squares:
\begin{equation}\label{eq:RSS}
RSS(k) = \min_{\mathbf{A},\mathbf{B}} ||\mathbf{X}-\mathbf{ABX}||^2
\end{equation}
A probabilistic formulation of linear AA is provided by \cite{seth2016} where it is observed that AA follows a simplex latent variable model and normal observation model. The generative process for the observations $\mathbf{x}_i$ in the presence of $k$ archetypes with archetype weights $\textbf{a}_i$ is given by
\begin{equation}\label{eq:probAT_1}
\mathbf{a}_i \sim \text{Dir}_k(\boldsymbol{\alpha}) \quad \text{ }\wedge\text{ } \quad
\mathbf{x}_i \sim \mathcal{N}(\mathbf{a}_iZ,\,\epsilon^2 \mathbf{I}),
\end{equation}
with uniform concentration parameters $\alpha_j = \alpha$ for all $j$, and weights summing up to $\lVert\mathbf{a}_i\rVert_1=1$. 
That is, the observations $\mathbf{x}_i$ are distributed according to isotropic Gaussians with means $\boldsymbol{\mu}_i=\mathbf{a}_iZ$ and variance $\epsilon^2$. 




%

\subsection{A Biological Motivation for Archetypal Analysis}
\label{sec:BiologicalMotivation}
\begin{figure}
\centering
\includegraphics[width=0.40\textwidth]{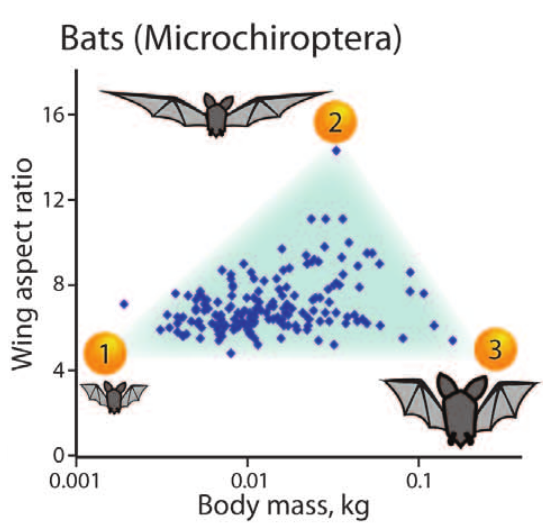}
\caption{Phenospace of different species of Microchiroptera. The dominant food habit of each species, and thereby the ability to procure this food source, is linked to the morphology of the animals, e.g. a higher Wing Aspect Ratio corresponds with the greater aerodynamic efficiency needed to chase high flying insects. Archetypes are extreme types, optimized to perform a single task. Proximity of a species to an archetype quantifies the level of adaptation this species has undergone with respect to the optimization objective or task. Reprinted from \cite{shoval2012} with permission.}
\label{fig:bats138}
\end{figure}

Conceptionally, the motivation for Archetypal Analysis is purely statistical but the method itself always implied the possibility of interpretations with a more \textit{evolutionary flavour}. By representing an individual data point as a mixture of \textit{pure types} or \textit{archetypes}, a natural link to the evolutionary development of biological systems is implicitly established. The publication by \cite{shoval2012} entitled 'Evolutionary Trade-Offs, Pareto Optimality, and the Geometry of Phenotype Space' made this connection explicit, providing a theoretical foundation of the 'archetype concept'. In general, evolutionary processes are multi-objective optimization problems and as such subject to unavoidable trade-offs: If multiple tasks need to be performed, no (biological) system can be optimal at all tasks at once. Examples of such trade-offs include those between longevity and fecundity in Drosophila melanogaster where long-lived flies show decreased fecundity \citep{drosophila} or predators that evolve to be fast runners but eventually have to trade-off their ability to subdue large or strong prey, e.g. cheetah versus lion \citep{garland2014}. Such evolutionary trade-offs are known to affect the range of phenotypes found in nature \citep{tendler2015}. In \cite{shoval2012} it is argued that best-trade-off phenotypes are weighted averages of archetypes while archetypes themselves are phenotypes specialized at performing a \textit{single} task optimally. An example of an evolutionary trade-off in the space of traits (or phenospace) for different species of bats (Microchiroptera) is shown in Figure \ref{fig:bats138}. Based on a study of bat wings by \cite{norberg1987}, each species is represented in a two-dimensional space where the axis depict Body Mass and Wing Aspect Ratio. The latter is the square of the wingspan divided by the wing area.
%
%
%
%
Table \ref{tab:bats138} gives an account of the task the archetypes indicated in Figure \ref{fig:bats138} have evolved to performing optimally. The trade-off situation can be interpreted using Pareto optimality theory \citep{steuer1986}, which was recently used in biology to study trade-offs in evolution \citep{schuetz2012,elSamad2005}. All phenotypes that have evolved over time lie within a restricted part of the phenospace, the so-called Pareto front, which is the set of phenotypes that cannot be improved at all tasks simultaneously. If there were a phenotype being better at all tasks than a second phenotype, then the latter would be eliminated over time by natural selection. Consequently phenotypes on the Pareto front are the best possible compromise between the different requirements or tasks.
%
%
\begin{center}
\begin{table}[ht]
\begin{tabular}{ccc}
\toprule
\thead{Archetype} & \thead{Phenotype} & \thead{Specialization} \\
\midrule
\rowcolor{gray!30}  
1 &  \makecell{\hspace{.2em}low aspect ratio,\hspace{.2em} \\ small body}  & \makecell{hunting small insects \\ near vegetation} \\
\rowcolor{white}
2 &  \makecell{high aspect ratio, \\ medium body}  & \makecell{hunting high flying \\ large insects} \\
\rowcolor{gray!30}
3 &  \makecell{\hspace{.2em}low aspect ratio,\hspace{.2em} \\ large body}  & \makecell{\hspace{1.0em}hunting animals\hspace{1.0em} \\ near vegetation} \\
\bottomrule
\end{tabular}
\caption{Inferred specialization of the archetypal species of Microchiroptera indicated in Figure \ref{fig:bats138}. From an evolutionary perspective, the phenotype is a consequence of the specialization, for details see \citep{shoval2012}.}
\label{tab:bats138}
\end{table}
\end{center}
%
%
\section{Related Work}
Linear ``Archetypal Analysis'' (AA) was first proposed by  
\cite{cutlerBreiman1994}. Since its conception, AA has known several advancements on the \textit{algorithmic side}: In \cite{spatiotempAT} the authors propose an archetype model able to identify archetypes in space \textit{and} time, named \enquote{archetypal analysis of spatio-temporal dynamics}. A similar problem is addressed in \enquote{Moving archetypes} by \cite{movingAT}. Model selection is the topic of \cite{sand2012}, where the authors are concerned with the optimal number of archetypes needed to characterize a given data set. An extension of the original archetypal analysis model to non-linear kernel archetypal analysis is proposed by \cite{bauck2014Kernel,morupKernelAA}. In \cite{kauf2015}, the authors use a copula based approach to make AA independent  of strictly monotone transformations of the input data. The reasoning is that such transformations should in general not influence which points are identified as archetypes. Algorithmic improvements by adapting a Frank--Wolfe type algorithm to speed-up the calculation of archetypes are made by \cite{bauck2015FW}. A probabilistic version of archetypal analysis was introduced by \cite{seth2016}, lifting the restriction of archetypal analysis to real--valued data and instead allowing other observation types such as integers, binary, and probability vectors as input. Efficient \enquote{coresets for archetypal analysis} are proposed by \cite{at-coresets} in order reduce the high computational cost due to the additional convexity-preserving constraints when identifying archetypes.

Although AA did not prevail as a commodity tool for pattern analysis, several applications have used it very successfully. In \cite{chan2003}, AA is used to analyse galaxy spectra which are viewed as weighted superpositions of the emissions from stellar populations, nebular emissions and nuclear activity. For the human genotype data studied by \cite{huggins2007}, inferred archetypes are interpreted as representative populations for the measured genotypes. In computer vision, AA has for example been used by \cite{bauck2009ImgCol} to find archetypal images in large image collections or by \cite{can2015} to perform the analogous task for large document collections. In combination with deep learning, archetypal style analysis \citep{styleAA} applies AA to learned image representations in order to realize artistic style manipulations.

Our work is based on the variational autoencoder model (VAE), arguably one of the most popular representatives of the class of ``Deep Latent Variable Models''. VAEs were introduced by \cite{kingmaWelling2013,rezende2014} and use an inference network to perform a variational approximation of the posterior distribution of the latent variables. Important work in this direction include \cite{KingmaSemi,pmlrrezende15} and \cite{Jang}. More recently, \cite{alemi2016} have discovered a close connection between VAE models and the Information Bottleneck principle \citep{tishby2000}. Here, the Deep Variational Information Bottleneck (DVIB) is a VAE where not the input $X$ is reconstructed (i. e. decoded) but rather a datum $Y$, about which $X$ is known to contain information. Subsequently, the DVIB has been extended in multiple directions such as sparsity \citep{2018arXiv180406216W} or causality \citep{2018arXiv180702326P}.\\
Akin to our work, \textit{AAnet} is a model proposed by \cite{AAnet} as an extension of linear archetypal analysis on the basis of standard, i. e. non-variational, autoencoders. In their work two regularization terms, applied to an intermediate representation, provide the latent archetypal convex representation of a non-linear transformation of the input. In contrast to our work, which is based on probabilistic generative models (VAE, DVIB), \textit{AAnet} attempts to emulate the generative process by adding noise to the latent representation during training. Further, no side information is incorporated which can -- and in our opinion should -- be used to constrain potentially \textit{over-flexible} neural networks and guide the optimisation process towards learning a meaningful representation. 
%
%
%
\section{Present Work}
Archetypal analysis, as proposed by \cite{cutlerBreiman1994}, is a linear method and cannot integrate any additional information about the data, e.g. labels, that might be available. Furthermore, the feature space in which AA is performed is spanned by features that had to be selected by the user based on prior knowledge. In the present work an extension of the original model is proposed such that appropriate representations can be learned end-to-end, side information can be incorporated to help learn these representations and non-linear relationships between features can be accounted for.


\subsection{Deep Variational Information Bottleneck}
We propose a model to generalise linear AA to the non-linear case based on the Deep Variational Information Bottleneck framework since it allows to incorporate side information $Y$ by design and is known to be equivalent to the VAE in the case of $Y=X$, as shown in \cite{alemi2016}. In contrast to the data matrix $X$ in linear AA, a non-linear transformation $f(X)$ giving rise to a latent representation $T$ of the data suitable for (non-linear) archetypal analysis is considered. I.e. the latent representation $T$ takes the role of the data $X$ in the previous treatment.\\
The DVIB combines the information bottleneck (IB) with the VAE approach \citep{tishby2000,kingmaWelling2013}. The objective of the IB method is to find a random variable $T$ which, while compressing a given random vector $X$, preserves as much information about a second given random vector $Y$. The objective function of the IB is as follows
\begin{equation}
\mbox{min}_{p(\mathbf{t}|\mathbf{x})} I(X;T) - \lambda I(T;Y),
\label{eq:ib1}
\end{equation}
where $\lambda$ is a Lagrange multiplier and $I$ denotes the mutual information. Assuming the IB Markov chain $T-X-Y$ and a parametric form of Eq. \ref{eq:ib1} with parametric conditionals $p_\phi(\mathbf{t}|\mathbf{x})$ and $p_\theta(\mathbf{y}|\mathbf{t})$, Eq. \ref{eq:ib1} is written as
\begin{equation}
\max_{\phi,\theta}  -I_{\phi}(\mathbf{t};\mathbf{x}) + \lambda  I_{\phi,\theta}(\mathbf{t};\mathbf{y}).
\label{eq:ib_parametricForm}
\end{equation}
As derived in \cite{2018arXiv180406216W}, the two terms in Eq. \ref{eq:ib_parametricForm} have the following forms:
\begin{equation}
\label{eq:encoder_parametric}
\begin{aligned}
I_{\phi}(T;X) &= D_{KL}\left( p_\phi(\mathbf{t}|\mathbf{x}) p(\mathbf{x}) \| p(\mathbf{t}) p(\mathbf{x})\right) \\
&= \mathbb{E}_{p(\mathbf{x})}  D_{KL}\left(p_\phi(\mathbf{t}|\mathbf{x})\| p(\mathbf{t}) \right)
\end{aligned}
\end{equation}
and
\begin{equation}
\label{eq:decoder_parametric}
\begin{aligned}
I_{\phi,\theta}(T;Y) &=\ D_{KL}\left(\left[\int p(\mathbf{t}|\mathbf{y},\mathbf{x})p(\mathbf{y},\mathbf{x})\, \mathrm{d}\mathbf{x} \right] \| p(\mathbf{t}) p(\mathbf{y})\right)\\
&=\ \mathbb{E}_{p(\mathbf{x},\mathbf{y})} \mathbb{E}_{p_\phi(\mathbf{t}|\mathbf{x})}\log p_\theta(\mathbf{y}|\mathbf{t}) + h(Y).
\end{aligned}
\end{equation}
Here $h(Y)=-\mathbb{E}_{p(\mathbf{y})}\log p(\mathbf{y})$ denotes the entropy of $Y$ in the discrete case or the differential entropy in the continuous case. The models in Eq. \ref{eq:encoder_parametric} and Eq. \ref{eq:decoder_parametric} can be viewed as the encoder and decoder, respectively. Assuming a standard prior of the form $p(\mathbf{t})=\mathcal{N}(\mathbf{t};0,I)$ and a Gaussian distribution for the posterior $p_\phi(\mathbf{t}|\mathbf{x})$, the KL divergence in Eq. \ref{eq:encoder_parametric} becomes a KL divergence between two Gaussian distributions which can be expressed in analytical form as in \cite{kingmaWelling2013}. $I_\phi(T;X)$ can then be estimated on mini-batches of size $m$ as 
\begin{equation}
\label{eq:vae_encoder}
I_\phi(\mathbf{t};\mathbf{x}) \approx \frac1m \sum_i D_{KL}\left(p_\phi(\mathbf{t}|\mathbf{x}_i)\| p(\mathbf{t}) \right).
\end{equation}
As for the decoder, $\mathbb{E}_{p(\mathbf{x},\mathbf{y})} \mathbb{E}_{p_\phi(\mathbf{t}|\mathbf{x})}\log p_\theta(\mathbf{y}|\mathbf{t})$ in Eq. \ref{eq:decoder_parametric} is estimated using the reparametrisation trick proposed by  \cite{kingmaWelling2013,rezende2014}: 
\begin{equation}
\label{eq:vae_decoder}
\begin{aligned}
I_{\phi,\theta}(\mathbf{t};\mathbf{y}) &= \mathbb{E}_{p(\mathbf{x},\mathbf{y})} \mathbb{E}_{\boldsymbol{\varepsilon} \sim \mathcal{N}(0,I)} \sum_i  \log  p_{\theta}\left(\mathbf{y}_i|\mathbf{t}_i \right) \\ 
&\hspace{4mm}+\mbox{const.}
\end{aligned}
\end{equation}
with the reparametrisation
\begin{equation}
\mathbf{t}_i = \boldsymbol{\mu}_{i}(\mathbf{x}) + \text{diag}\left(\boldsymbol{\sigma}_i(\mathbf{x})\right) \boldsymbol{\varepsilon}.
\end{equation}
As mentioned earlier, in the case of $Y=X$ the original VAE is retrieved \citep{alemi2016}. In our applications, we would like to predict not only the side information $Y$ but also reconstruct the input $X$. Similar to the approach proposed in \cite{Bombarelli}, we use an additional decoder branch to predict the reconstruction $\Tilde{X}$. This extension requires an additional term $I_{\phi,\psi}(\mathbf{t};\mathbf{\Tilde{x}})$ in the objective function Eq. \ref{eq:ib_parametricForm} and an additional Lagrange multiplier $\nu$. The mutual information estimate $I_{\phi,\psi}(\mathbf{t};\mathbf{\Tilde{x}})$ is obtained analogously to Eq. \ref{eq:vae_decoder}.  
%
\subsection{Deep Archetypal Analysis}
\begin{figure}[h]
\centering
\includegraphics[width=0.50\textwidth]{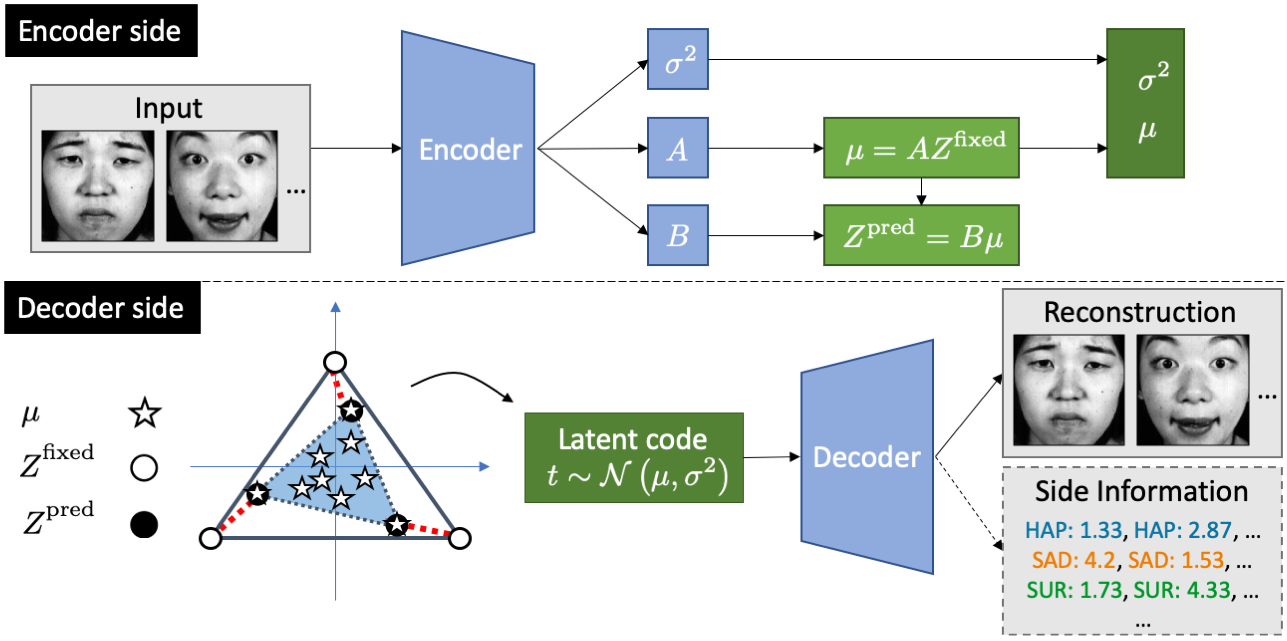}
\caption{Illustration of the \textit{Deep AA} model. \textbf{Encoder side}: Learning weight matrices $A$ and $B$ allows to compute the archetype loss $\ell_{AT}$ in Eq. \ref{eq:lossAT} and sample latent variables $\mathbf{t}$ as described in Eq. \ref{eq:deepAA_t}. The constraints on $A$ and $B$ in Eq. \ref{eq:constraint_A_B} are enforced by using softmax layers. \textbf{Decoder side}: $Z^\text{fixed}$ represent the fixed archetype positions in latent space while $Z^\text{pred}$ are given by the convex hull of the transformed data point means $\mu$ during training. Minimizing $\ell_{AT}$ corresponds to minimizing the red-dashed (pairwise) distances. The input is reconstructed from the latent variable $\mathbf{t}$. In the presence of side information, the latent representation allows to reproduce the side information $Y$ as well as the input $X$.}
\label{fig:at-supervised-arch}
\end{figure}

Deep Archetypal Analysis can then be formulated in the following way. For the sampling of $\mathbf{t}_i$ in Eq. \ref{eq:vae_decoder} the probabilistic AA approach as in Eq. \ref{eq:probAT_1} can be used which leads to
\begin{equation}
\label{eq:deepAA_t}
\mathbf{t}_i \sim \mathcal{N}\left(\boldsymbol{\mu}_i(\mathbf{x})=\mathbf{a}_i(\mathbf{x}) Z,\,\boldsymbol{\sigma}_i^2(\mathbf{x}) \mathbf{I}\right),
\end{equation}
where the mean $\boldsymbol{\mu}_i$ given through $\mathbf{a}_i$ and variance $\boldsymbol{\sigma}_i^2$ are non-linear transformations of the data point $\mathbf{x}_i$ learned by the encoder. We note that the means $\boldsymbol{\mu}_i$ are convex combinations of weight vectors $\mathbf{a}_i$ and the archetypes $\mathbf{z}_{j=1..k}$ which in return are considered to be convex combinations of the means $\boldsymbol{\mu}_{i=1..m}$ and weight vectors $\mathbf{b}_j$.\footnote{Note that $i=1..m$ (and not up to $n$), which reflects that deep neural networks usually require batch-wise training with batch size $m$.} By learning weight matrices $A\in \mathbb{R}^{m\times k}$ and $B\in \mathbb{R}^{k\times m}$ which are subject to the constraints formulated in Eq. \ref{eq:constraint_A_B} and parameterised by $\phi$, a non-linear transformation of data $X$ is learned which drives the structure of the latent space to form archetypes whose convex combination yield the transformed data points. A major difference to linear AA is that for \textit{Deep AA} we cannot identify the positions of the archetypes $\mathbf{z}_j$ as there is no absolute frame of reference in latent space. We thus position $k$ archetypes at the vertex points of a $(k-1)$-simplex and collect these \textit{fixed} coordinates in the matrix $Z^{\text{fixed}}$. These requirements lead to an additional distance-dependent archetype loss of
\begin{equation}\label{eq:lossAT}
\ell_{\text{AT}} = ||Z^{\text{fixed}}-BAZ^{\text{fixed}}||_2^2 = ||Z^{\text{fixed}}-Z^{\text{pred}}||_2^2,
\end{equation}
where $Z^\text{pred} = BAZ^{\text{fixed}}$ are the \textit{predicted} archetype positions given the learned weight matrices $A$ and $B$. For $Z^\text{pred}\approx Z^\text{fixed}$ the loss function $\ell_{\text{AT}}$ is minimized and the desired archetypal structure is achieved. The objective function of \textit{Deep AA} is then given by 
%
%
\begin{equation}
\max_{\phi,\theta}  -I_{\phi}(\mathbf{t};\mathbf{x}) + \lambda  I_{\phi,\theta}(\mathbf{t};\mathbf{y}) 
+ \nu  I_{\phi,\psi}(\mathbf{t};\mathbf{\Tilde{x}})-\ell_{\text{AT}}.
\label{eq:ib_DeepAA}
\end{equation}
A visual illustration of \textit{Deep AA} is given in Figure \ref{fig:at-supervised-arch}. The constraints on $A$ and $B$ can be guaranteed by using softmax layers and \textit{Deep AA} can be trained with a standard stochastic gradient descent technique such as Adam \citep{KingmaB14}. Note that the model naturally allows to be relaxed to the VAE setting by omitting the side information term $\lambda  I_{\phi,\theta}(\mathbf{t};\mathbf{y})$ in Eq. \ref{eq:ib_DeepAA}.

%
%

\subsection{The Necessity for Side Information}
\label{subsec:Necessity_for_Side_Information}
The goal of Deep AA is to identify meaningful archetypes in latent space which will subsequently allow the informed exploration of the given data set. The \enquote{meaning} of an archetype, and thereby the associated interpretation, can be improved by providing so-called side information, i.e. information in addition to the input data. If the input datum is for example an image, additional information could simply be a scalar- or vector-valued label. Using richer side information, e.g. additional images, is possible, too. The fundamental idea is that information about what constitutes a \textit{typical} representative (in the archetypal sense) might not be information that is readily present in the input $X$ but dependent on -- or even defined by -- the side information. Taking a data set of car images as an example, what would be an \textit{archetypal} car? Certainly, the overall size of a car would be a good candidate, such that smaller sports cars and larger pick-ups might be identified as archetypes. But introducing the fuel consumption of each car as side information would put sports cars and pick-ups closer together in latent space, as both car types often consume above average quantities of fuel. In this way, side information guides the learning of a latent representation which is informative with respect to exactly the side information provided. Consequently, typicality is not a characteristic of the data solely, but a function of the provided side information. And the selection of appropriate side information can only be linked to the question the user of a deep AA model tries to answer.

\section{Experiments}
\label{sec:experiments}
\subsection{Archetypal Analysis: Dealing With Non-linearity}
\begin{figure}
  \begin{subfigure}[t]{.48\textwidth}
  \centering
    \includegraphics[width=.95\textwidth]{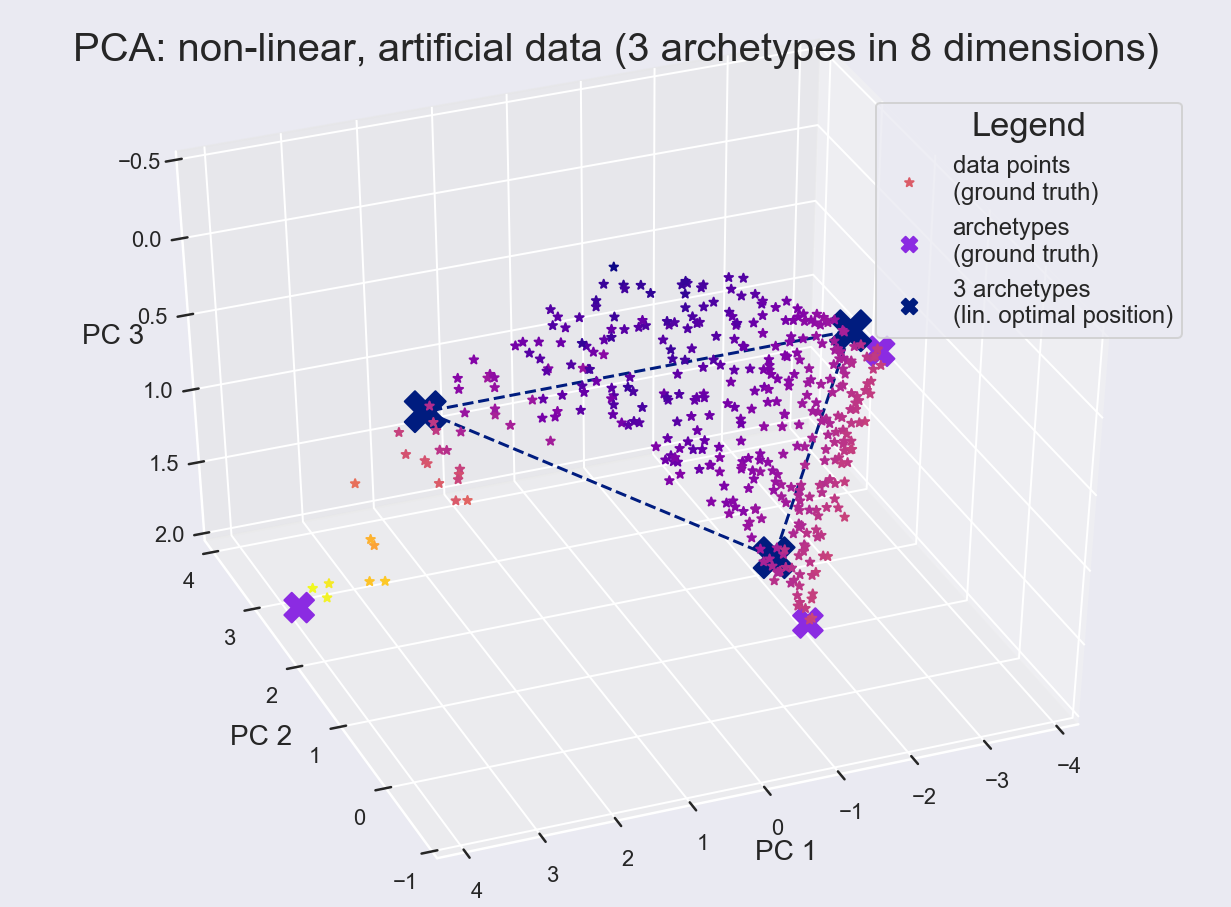}
    \caption{Linear AA is unable to recover the true archetypes.}
    \label{fig:linAA-vs-deepAA_panel:linAA}
  \end{subfigure}\hfill
  \begin{subfigure}[t]{.48\textwidth}
  \centering
    \includegraphics[width=.95\textwidth]{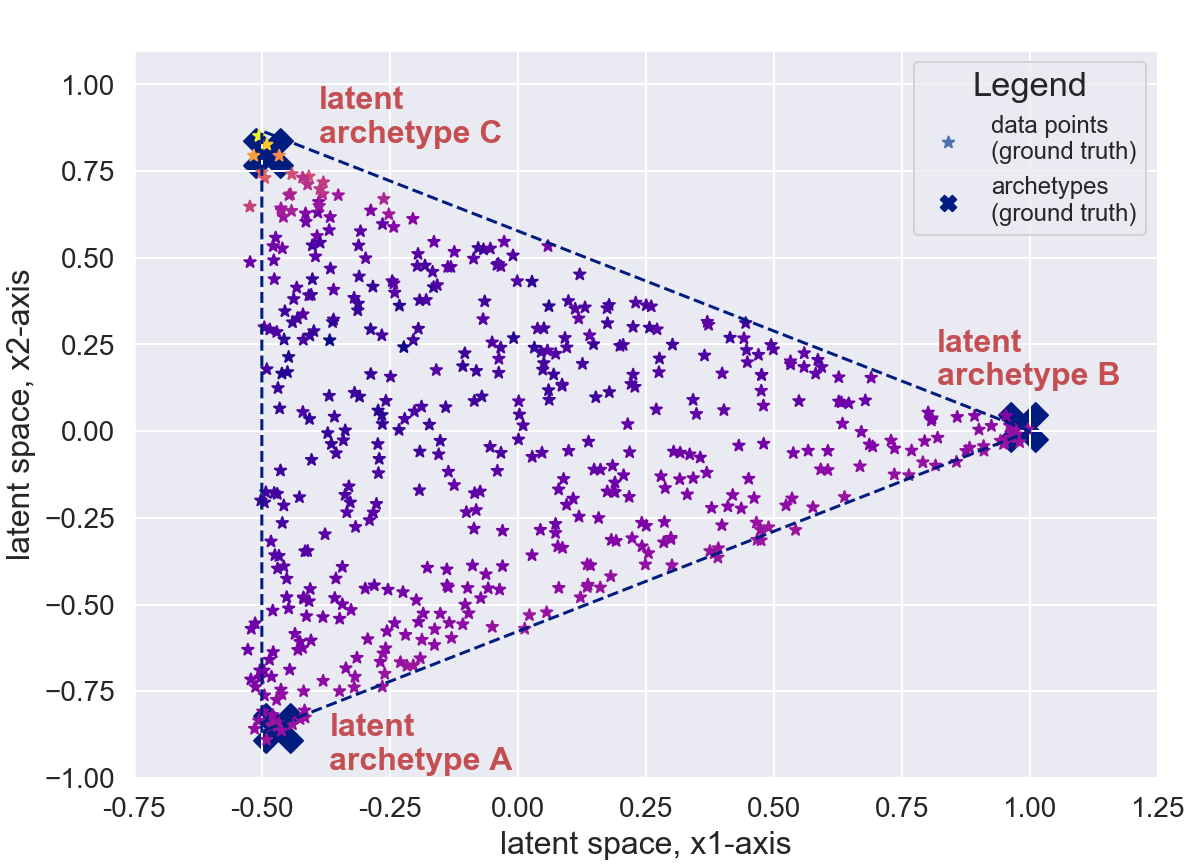}
    \caption{Latent space embedding of non-linear artificial data.}
    \label{fig:curvedSubmanifold5ATs}
  \end{subfigure}
  \caption{While linear archetypal analysis is in general unable to approximate the convex hull of a non-linear data set well, Deep AA learns an appropriate latent representation where the ground truth archetypes can correctly be identified.}
  \label{fig:linAA-vs-deepAA}
\end{figure}

\paragraph{Data generation.} For this experiment, data $\mathbf{X}\in \mathbb{R}^{n\times 8}$ is generated that is a convex mixture of $k$ archetypes $\mathbf{Z}\in \mathbb{R}^{k\times 8}$ with $k\ll n$. The generative process for the datum $\mathbf{x_i}$ follows Eq. \ref{eq:probAT_1}, where $\mathbf{a}_i$ is a stochastic weight vector denoting the fraction of each of the $k$ archetypes $\mathbf{z}_j$ needed to represent the data point $\mathbf{x}_i$. A total of $n=10000$ data points is generated, of which $k=3$ are true archetypes. The variance is set to $\sigma^2=0.05$ and the linear 3-dim data manifold is embedded in a $n=8$ dimensional space. Note that although linear and deep archetypal analysis is always performed on the full data set, only a fraction of that data is displayed when visualizing results.
\paragraph{Linear AA -- non-linear data.} Data is generated as described above and an additional non-linearity is introduced by applying an exponential to one dimension of $\mathbf{X}$ which results in a curved 8-dimensional data manifold. Linear archetypal analysis is then performed using the efficient Frank-Wolfe procedure proposed by \cite{bauck2015FW}. For visualization, PCA is used to recover the original 3-dimensional data submanifold which is embedded in the 8-dimensional space. The first three principal components of the ground truth data are shown in Figure \ref{fig:linAA-vs-deepAA_panel:linAA} as well as the computed archetypes (connected by dashed lines). The positions of the computed archetypes occupy optimal positions according to the optimization problem in equation \ref{eq:RSS} but due to the non-linearity in the data it is impossible to recover the three ground truth archetypes.
\paragraph{Deep AA -- non-linear data.} For data that has been generated as described in the previous paragraph, a strictly monotone transformation in form of an exponentiation should in general \textit{not} change which data points are identified as archetypes. But this is clearly the case for linear AA as it is unable to recover the true archetypes \textit{after} a non-linearity has been applied. Using that same data to train the deep AA architecture presented in Figure \ref{fig:at-supervised-arch} generates the latent space structure shown in Figure \ref{fig:curvedSubmanifold5ATs}, where the three archetypes A, B and C have been assigned to the appropriate vertices of the latent simplex. Moreover, the sequence of color stripes shown has been correctly mapped into the latent space. Within the latent space data points are again described as convex linear combinations of the latent archetypes. Latent data points can also be reconstructed in the original data space through the learned decoder network. The network architecture used for this experiment was a simple feedforward network (2 layered encoder and decoder), training for 20 epochs with a batch size of 100 and a learning rate of 0.001.

\subsection{Archetypes in Image-based Sentiment Analysis}
The Japanese Female Facial Expression (JAFFE) data\-base was introduced by \citet{lyons1998} and contains 213 images of 7 facial expressions (6 basic facial expressions + 1 neutral). The expressions are happiness, sadness, surprise, anger, disgust and fear. All expressions were posed by 10 Japanese female models. Each image has been rated on 6 emotion adjectives by 60 Japanese subjects on a 5 level scale (5-high, 1-low) and each image was then assigned a 6-dim. vector of average ratings. For the following experiments the advice of the creator of the JAFFE data set was followed to exclude \textit{fear} images and the \textit{fear} adjective from the ratings, as the models were not believed to be good at posing fear. All experiments based on the JAFFE data set are performed on the following architecture\footnote{The code is available via \url{https://github.com/bmda-unibas/DeepArchetypeAnalysis}}:

\begin{description}
    \item \textbf{Encoder}: \\
    Input: image $\mathbf{x}$ (128$\times$128)
    \\$\rightarrow$ 3$\times$\Big[64 Conv. (4$\times$4) + Max-Pool. (2$\times$2)\Big] \\
    $\rightarrow$ Flatten + FC100 \\ 
    $\rightarrow$ $\mathbf{A}$, $\mathbf{B}$, $\sigma^2$
    \item \textbf{Decoder (Image Branch)}: \\
    Input: latent code $\mathbf{t}$\\
    $\rightarrow$ FC49 \\
    $\rightarrow$ 3$\times$\Big[64 Conv. Transpose (4$\times$4)\Big] \\
    $\rightarrow$ Flatten + FC128$\times$128 \\
    $\rightarrow$ FC128$\times$128 $\rightarrow$ 128$\times$128 reconstruction $\mathbf{\Tilde{x}}$
    \item \textbf{Decoder (Side Information Branch)}: \\
    Input: latent code $\mathbf{t}$  \\$\rightarrow$ FC200-5 $\rightarrow$ side information $\mathbf{\Tilde{y}}$
\end{description}

ReLU activations are used in-between layers and sigmoid activations for the image intensities.
The different losses are weighted as follows: we multiplied the archetype loss by a factor of 80,  the side information loss by 560, and the KL divergence by 40. 
In the setting where only two labels are considered, the weight for archetype loss is increased to 120. 
The network was trained for 5000 epochs with a mini-batch size of 50 and a learning rate of 0.0001.
For training a NVIDIA TITAN X Pascal GPU was used, where a full training sessions lasted approximately 30 minutes.

\subsubsection{JAFFE: Latent Space Structure}
\begin{figure}[h!]
\centering
\includegraphics[width=0.50\textwidth]{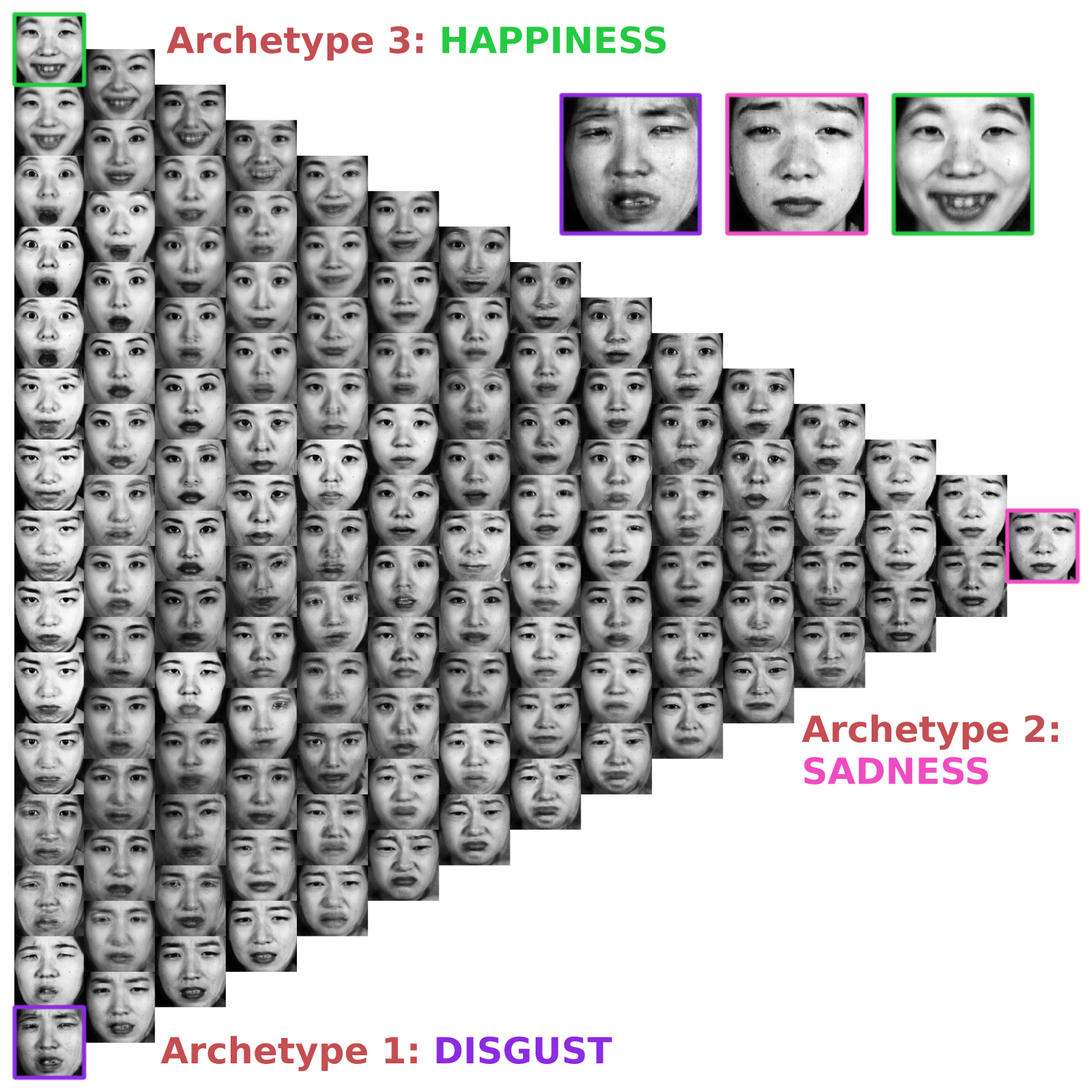}
\caption{Latent structure of the JAFFE data set when trained on a subset of the side information containing only the emotion ratings for \enquote{sadness} and \enquote{disgust}.}
\label{fig:jaffe_LatentStructureOfJAFFE}
\end{figure}

Emotions conveyed through facial expressions are a suitable case to demonstrate the interpretability of learned latent representation in deep AA. First, the existence of archetypes is plausible as there clearly are expressions that convey a maximum of a given emotion, i. e. a person can look extremely/maximally surprised.
%
%
\begin{figure}[h!]
  \begin{subfigure}[t]{.50\textwidth}
  \centering
    \includegraphics[width=.95\textwidth]{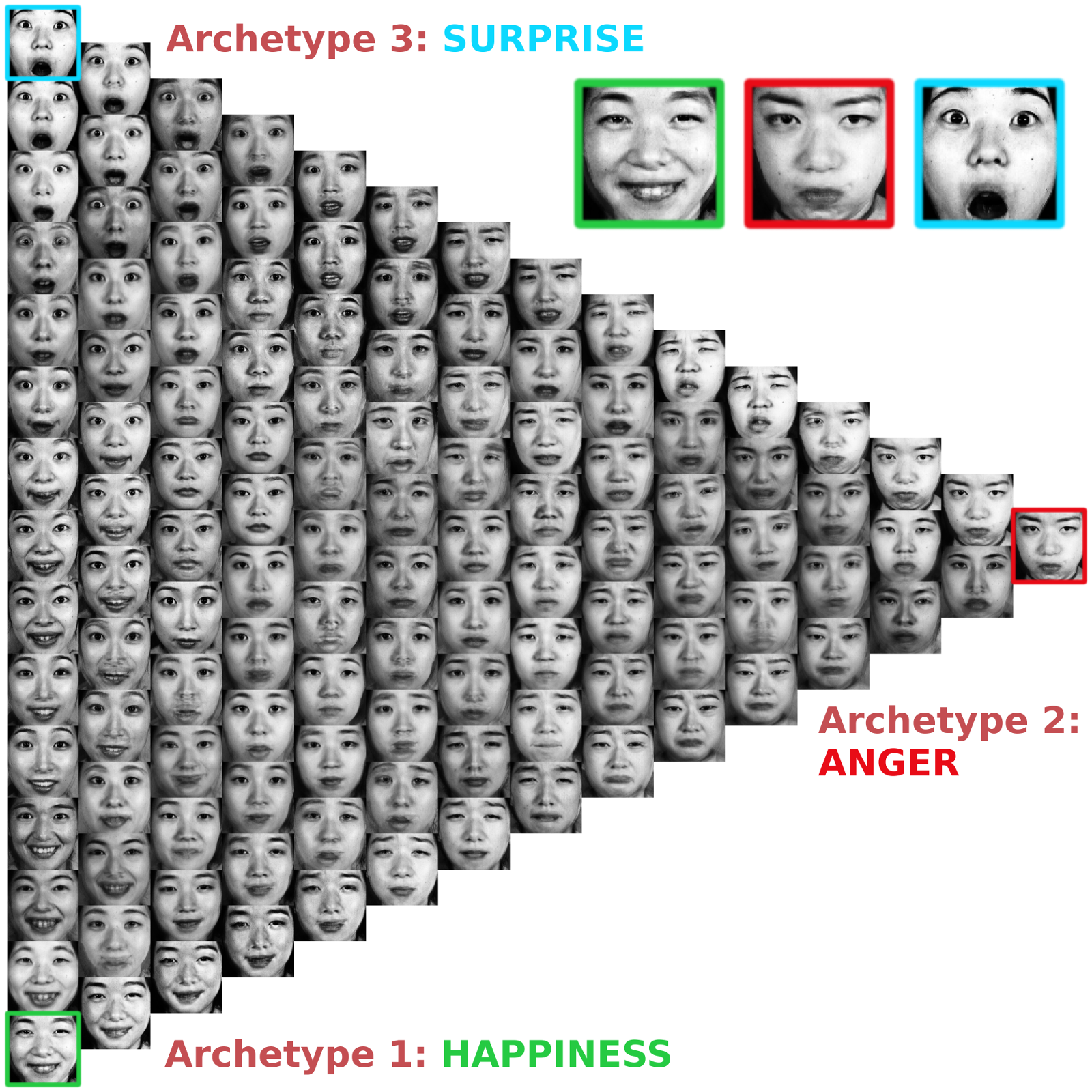}
    \caption{Archetype latent space of the JAFFE data set.}
    \label{fig:jaffe_latenImages}
  \end{subfigure}\hfill
  \begin{subfigure}[t]{.50\textwidth}
  \centering
    \includegraphics[width=.95\textwidth]{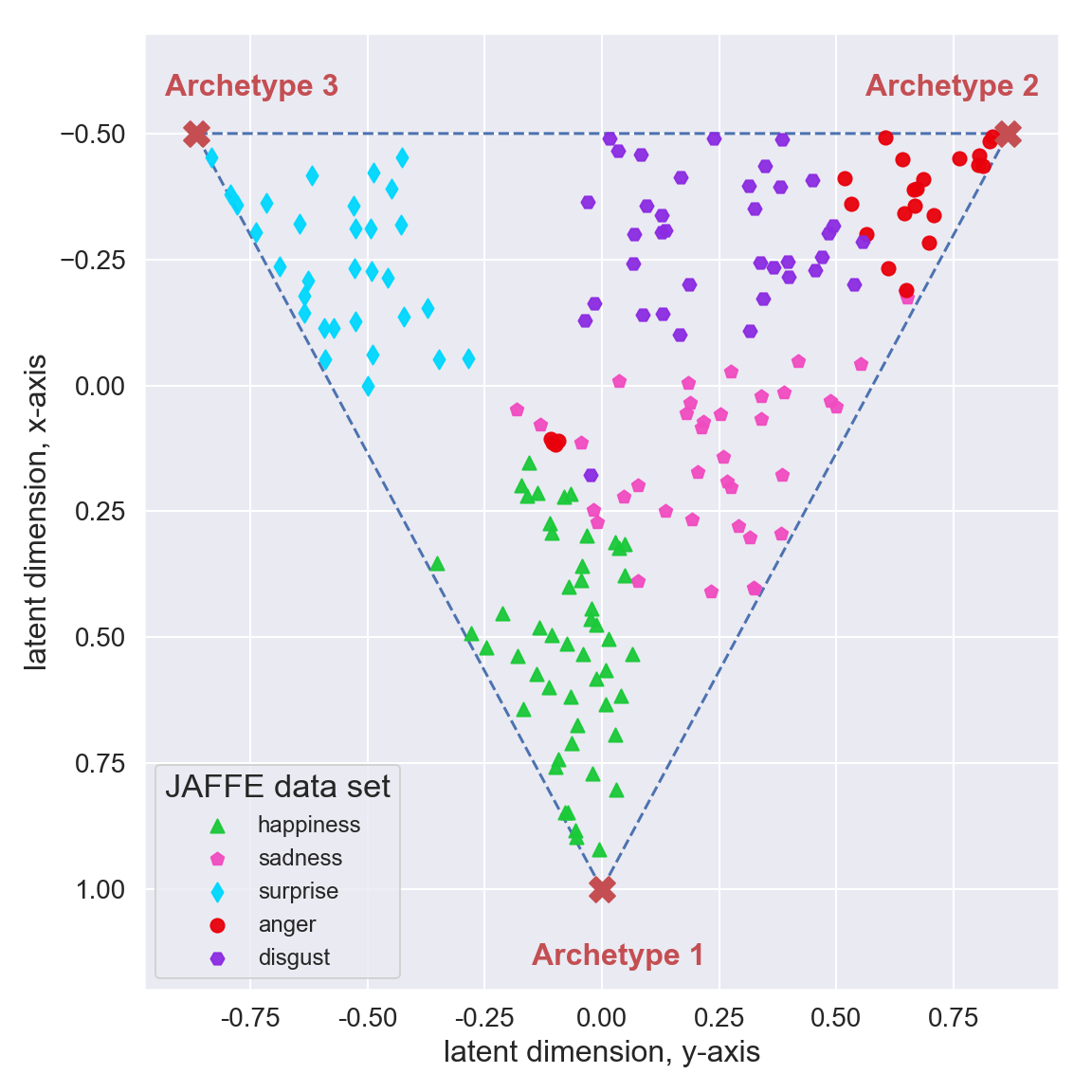}
    \caption{Location of emotion adjectives in latent space.}
    \label{fig:jaffe_latentMeans}
  \end{subfigure}
  \caption{Deep AA with $k=3$ archetypes identifies sadness as a mixture mostly between happiness and anger while disgust lies between the archetypes for anger and surprise.}
\end{figure}
%
%
Second, facial expressions change continuously without having a clearly defined cluster structure. Moreover, these expressions lend themselves to being interpreted as mixtures of basic (or archetypal) emotional expressions -- a perspective also enforced by the averaged ratings for each image which are essentially weight vectors with respect to the archetypal emotional expressions. Figure \ref{fig:jaffe_latenImages} shows the learned archetypes \enquote{happiness}, \enquote{anger} and \enquote{surprise} while expressions linked to the emotion adjective \enquote{sadness} are identified as mixtures between archetype 1 (happiness) and archetype 2 (anger). Figure \ref{fig:jaffe_latentMeans} shows the positions of the latent means where the color coding is based on the \textit{argmax} of the emotion rating, which is a 5-dimensional vector. An analogous situation is found in case of \enquote{disgust}, which, according to deep AA, is a mixture between archetype 2 (anger) and archetype 3 (surprise). Towards the center of the simplex, expressions are located which share equal weights with respect to the archetypes and thus resemble a more \enquote{neutral} facial expression, as shown in figure \ref{fig:jaffe_LatentStructureOfJAFFE}.
\paragraph{Side Information for JAFFE.} The JAFFE data set contains facial expressions posed by 10 Japanese female models. Based solely on the visual information, i.e. disregarding the emotion scores, these images could meaningfully be grouped together in a variety of ways, e. g. head shape, hair style, identity of the model posing the expressions... Interpreting resulting archetypes in general requires guiding information that tells the model which \enquote{definition of typicality} it is required to learn. While it is obvious to learn typical emotion expressions in case of JAFFE, most applications are arguably more ambiguous. In section \ref{chemEx} a chemical experiment is discussed, where each molecule can be described by a variety of properties. 
The side information introduced to the learning process will ultimately be the property the experimenter is interested in, and typicality will have to be understood with respect to that property.

\subsubsection{JAFFE: Expressions As Weighted Mixtures}
\begin{figure}[h!]
\centering
\includegraphics[width=0.50\textwidth]{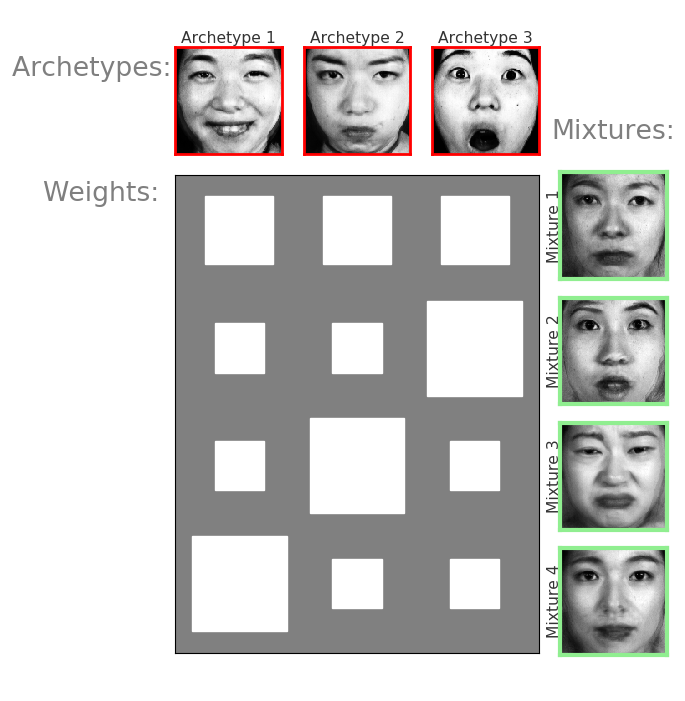}
\caption{Knowing the archetypes allows for an informed exploration of the latent space by \textit{not} directly sampling latent space coordinates but by specifying a desired mixture with respect to the known archetypes.}
\label{fig:jaffe_weightedMixture}
\end{figure}

One advantage of deep AA compared to the plain Variational Autoencoder (VAE) is a \textit{globally} interpretable latent structure. All latent means $\mu_i$ will be mapped inside the convex region spanned by the archetypes. And as archetypes represent extremes of the data set which are present to some percentage in all data points, these percentages or weights can be used to explore the latent space in an \textit{informed} fashion. This might be especially of advantage in case of higher-dimensional latent spaces. For example will the center of the simplex always accommodate latent representations of input data that are considered \textit{mean} samples of the data set. Moreover, directions within the simplex have meaning in the sense that when \enquote{walking} towards or away from a given archetype, the characteristics of that archetype will either be enforce or diminished in the decoded datum associated with the actual latent position. This is shown in the Hinton plot in Figure \ref{fig:jaffe_weightedMixture} where mixture 1 is a mean sample, i. e. with equal archetype weights. Starting at this position and moving on a straight line into the direction of archetype 3 increases its influence while equally diminishing the influence of both archetypes 1 and 2. This results in mixture 2 which starts to look surprised, but not as extremely surprised as archetype 3. In the same fashion mixture 3 and 4 are the results of walking straight into the direction of archetypes 2 or 1 which results in a sad face (mixture 3) and a slightly happy facial expression (mixture 4).      
%
%
%
%
\subsubsection{JAFFE: Deep AA Versus VAE}
\begin{figure}
\centering
\begin{subfigure}{0.45\textwidth}
   \includegraphics[width=1\linewidth]{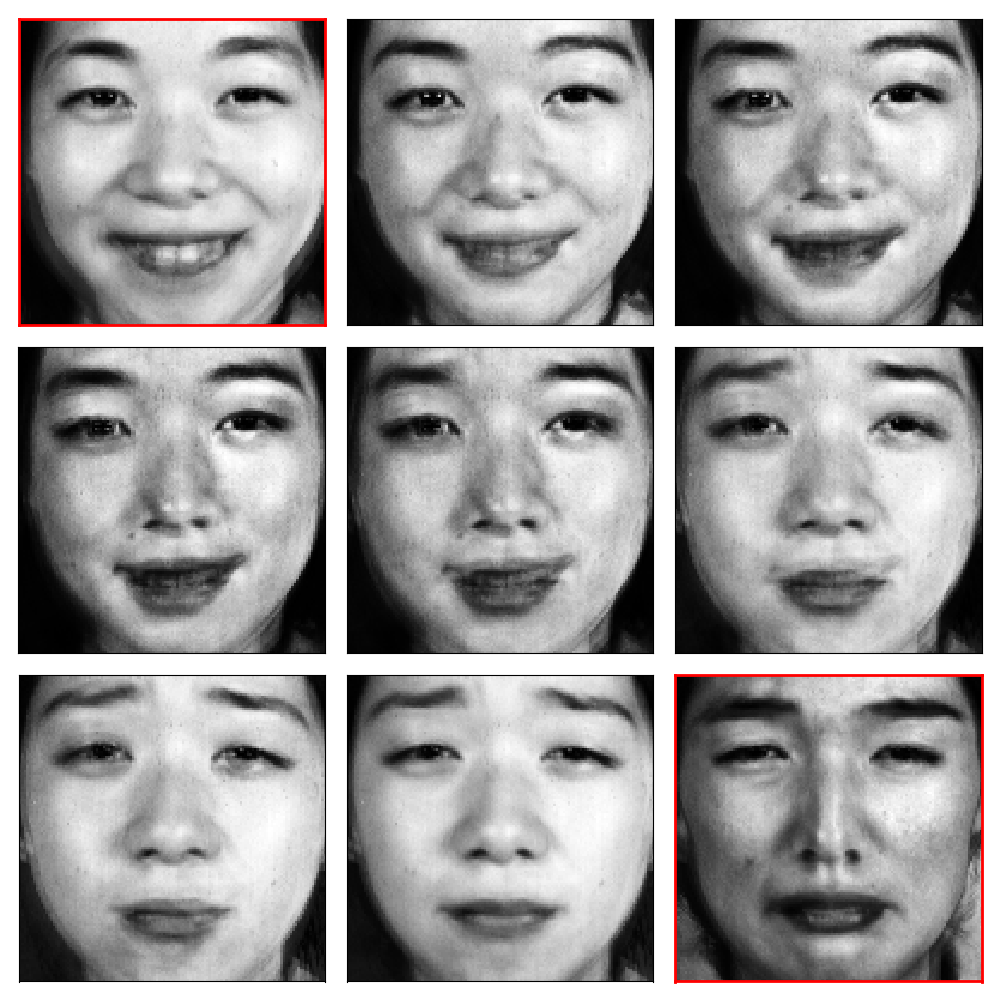}
   \caption{Interpolation based on deep AA}
   \label{fig:jaffe_DAA_interpolation} 
\end{subfigure}
\begin{subfigure}{0.45\textwidth}
   \includegraphics[width=1\linewidth]{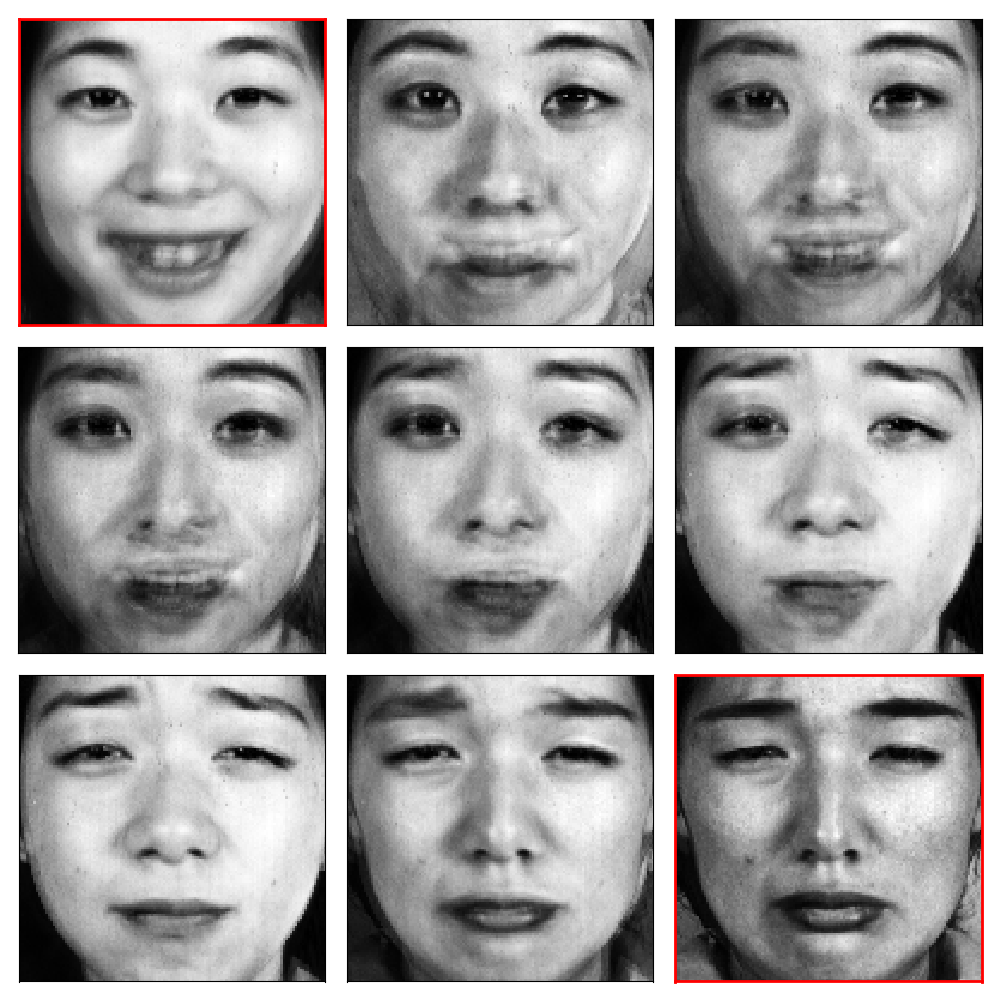}
   \caption{Interpolation based on VAE (with lin. AA)}
   \label{fig:jaffe_VAE_interpolation}
\end{subfigure}
\caption{The location of two input images were located in the latent space of the deep AA and the VAE model with subsequently performed linear AA. The interpolation is qualitatively better in case of deep AA where latent means are mapped more densely together due to the simplex constraints.}
\end{figure}

Deep AA is designed to be a model that simultaneously learns an appropriate representation and identifies meaningful latent archetypes. This model can be compared to a plain VAE where a latent space is learned first and subsequently linear AA is performed on that space in order to approximate the latent convex hull. Figure \ref{fig:jaffe_DAA_interpolation} shows the interpolation in the deep AA model between two images, neither of them archetypes, from \enquote{happy} to \enquote{sad}. Compared to Figure \ref{fig:jaffe_VAE_interpolation}, which shows the same interpolation in a VAE model with subsequently performed linear AA, the interpolation based on deep AA gives a markedly better visual impression. In case of deep AA, this is explained by the fact that all data points are mapped into the simplex which ensures a relatively dense distribution of the latent means. On the other hand, the latent space of the VAE model has no hard geometrical restrictions and thus the distribution of the latent representatives will be less dense or even \enquote{patchy}, i. e. with larger empty areas in latent space. Especially with small data sets such as JAFFE, of which less than 200 images are used, interpolation quality might be strongly affected by the unboundedness of the latent space of VAE models.
%
%
%
%
%
%
%
%

\subsection{The Chemical Universe Of Molecules}
\label{chemEx}
In the following section the application of deep AA to the domain of chemistry is explored. Starting with an initial set of chemical compounds, e. g. small organic molecules with cyclic cores \citep{visini2017}, and iteratively applying a finite number of reactions, will eventually lead to a huge collection of molecules with extreme combinatorial complexity. But while the total number of all possible \textit{small} organic molecules has been estimated to exceed $10^{60}$ \citep{kirkpatrick2004}, even this number pales in comparison to the whole chemical universe of organic chemistry. In general, the efficient exploration of chemical spaces requires methods capable of learning meaningful representations and endowing these spaces with a globally interpretable structure. Prominent examples of chemistry data sets include the family of GDB-xx data sets (generic database), e. g. GDB-13 \citep{gdb13}, which enumerates small organic molecules of up to 13 atoms, composed of the elements C, N, O, S and Cl, following simple chemical stability and synthetic feasibility rules. With more than 970 million structures, GDB-13 is the largest publicly available database of small organic molecule to date.
%
\begin{figure}[hb!]
  \includegraphics[width=0.5\textwidth]{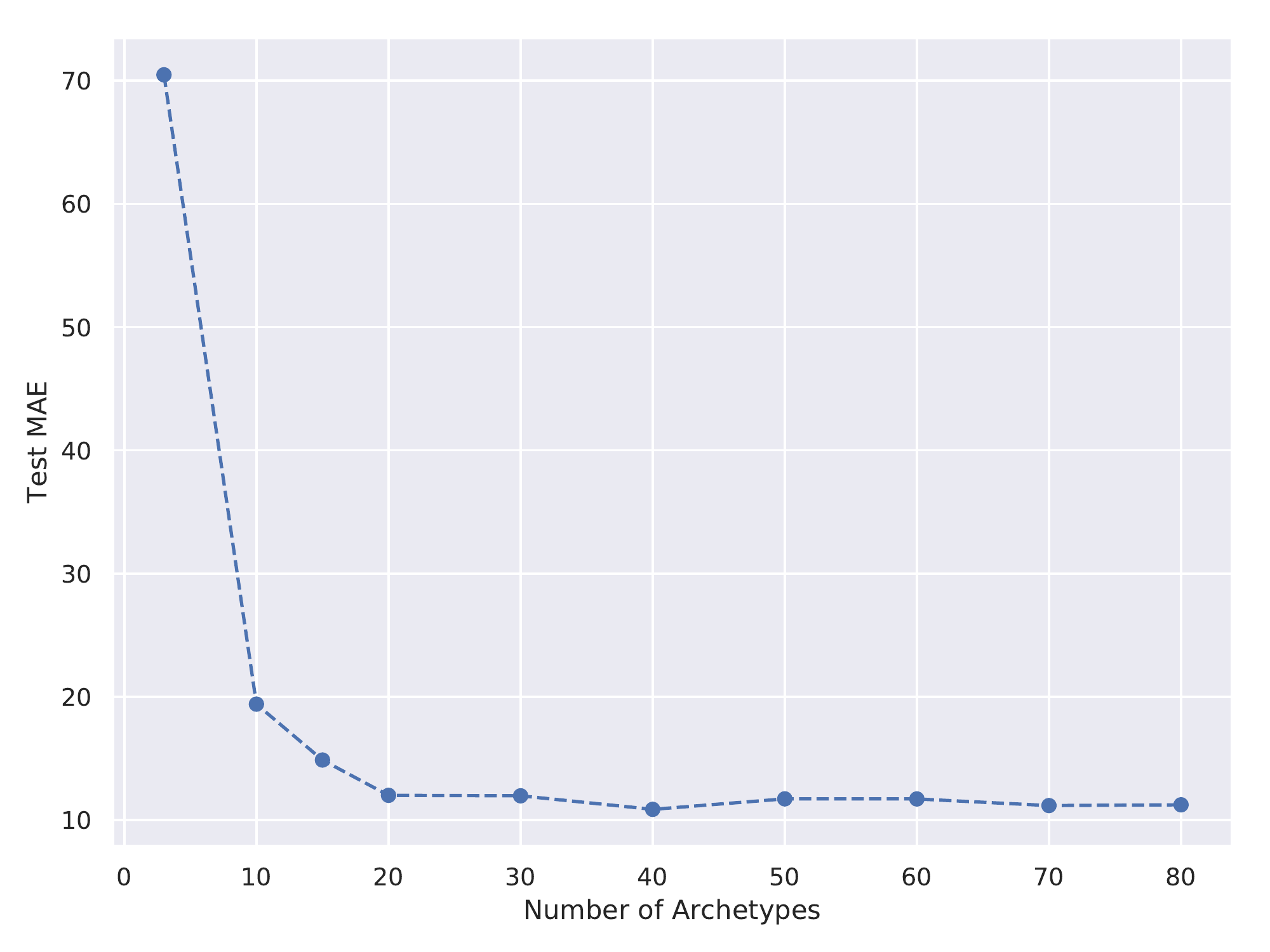}
\caption{Model selection on the QM9 data set: Mean absolute error (reconstruction loss) vs. number of archetypes on the test set.}
\label{fig:qm9selection}
\end{figure}
\begin{figure*}[!h]
    \centering
    \begin{subfigure}{0.45\textwidth}
        \centering
        \includegraphics[width=\textwidth]{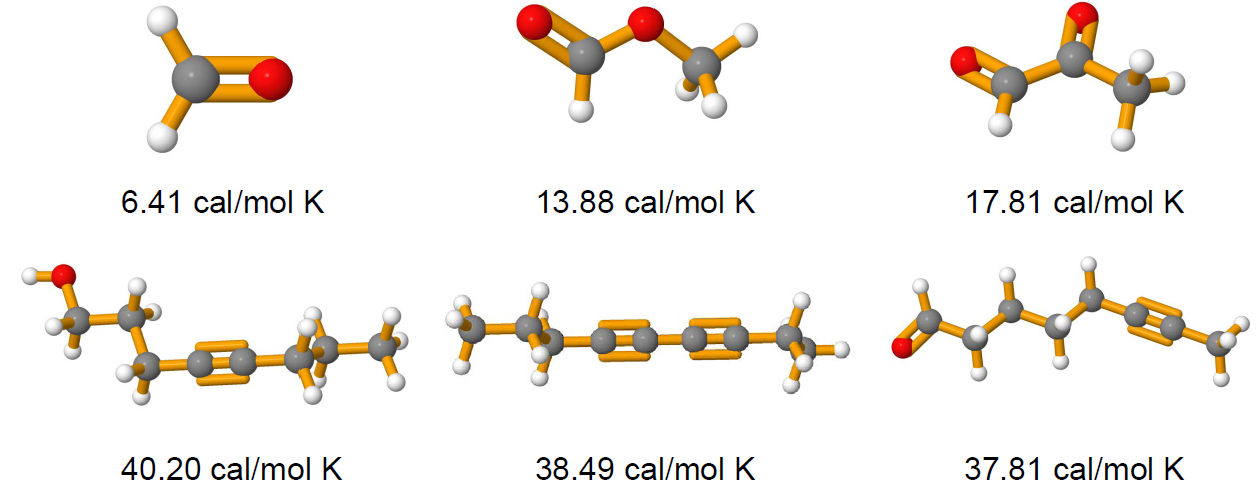}
        \caption{}\label{figure:case1}
    \end{subfigure} \hspace{1cm}
    \begin{subfigure}{0.45\textwidth}
        \centering
        \includegraphics[width=0.9\textwidth]{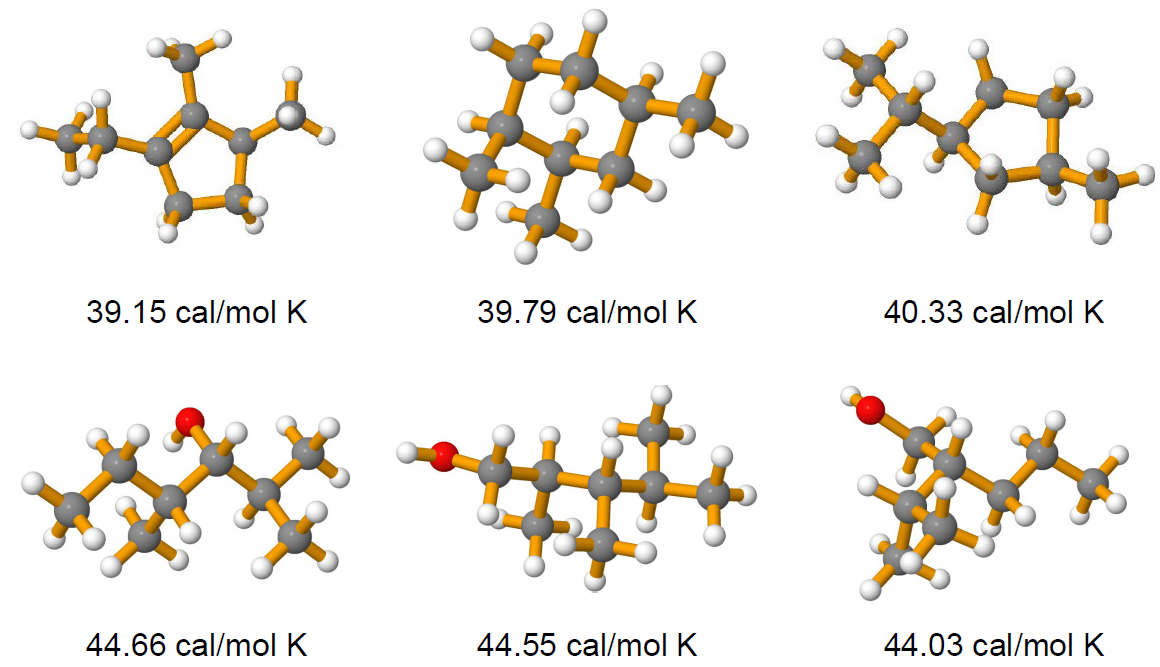}
        \caption{}\label{figure:case2}
    \end{subfigure}
    \caption{Both panels illustrate a comparison between archetypal molecules, where typicality is understood with respect to the molecular property \textit{heat capacity}. Each column contains the three molecules of the test set that have been mapped closest to a specific vertex of the latent simplex. Panel (a) compares archetypal \textit{linear} molecules characterized by a short chain structure versus long chained molecules. Panel (b) compares archetypal molecules with similar masses but different geometric configuration, i. e. with and without a cyclic structure.}
    \label{fig:comparisions}
\end{figure*}
\paragraph{Exploring The Chemical Space.} As discussed in section \ref{sec:BiologicalMotivation}, archetypal analysis lends itself to a distinctly evolutionary interpretation. Although this is certainly a more biological perspective, the basic principle is applicable to other fields. In chemistry, the principle of \textit{evolutive abiogenesis} describes a process in which simple organic compounds increase in complexity \citep{miller53}. In the following experiment a structured chemical space is learned using as side information the \textit{heat capacity} $C_v$ which quantifies the amount of energy (in Joule) needed to increase 1 Mol of molecules by 1 K at constant volume. A high $C_v$ number is important e. g. in applications dealing with the storage of thermal energy \citep{CABEZA20151106}. In the following, all experiments are based on the QM9 data set \citep{rama2014,rudd2012}, which contains molecular structures and properties of 134k organic molecules. Each molecule is made up of nine or less atoms, i. e.  C, O, N, or F, without counting hydrogen. The QM9 data set is based on ab-initio density functional theory (DFT) calculations.
%
%
%
\begin{figure*}[!h]
\begin{center}
  \includegraphics[width=1.0\textwidth]{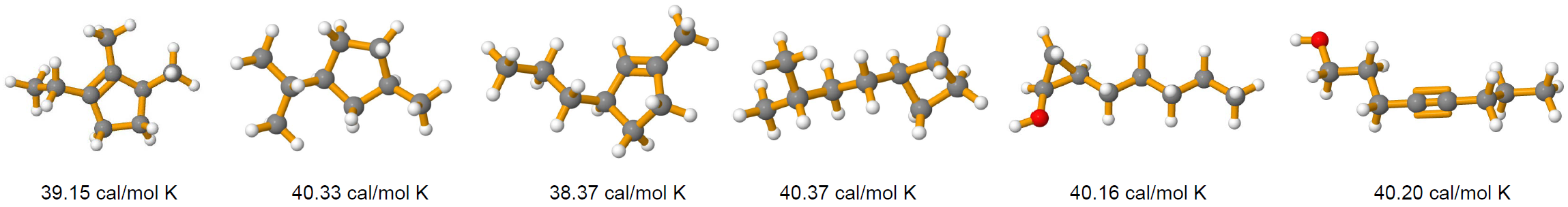}
\caption{Interpolation between two archetypal molecules produced by deep AA. The labels display the heat capacity of each molecule. Here, only a single example is shown but similar results can be observed for other combinations of archetypes.}
\label{fig:interpolation}
\end{center}
\end{figure*}
%
%
\paragraph{Experiment Setup.} A total of 204 features were extracted for every molecule using the Chemistry Development Kit \citep{cdk}. The neural architecture used has 3 hidden FC layers with 1024, 512 and 256 neurons, respectively and ReLU activation functions. For all experiments, the model was trained in a \textit{supervised} fashion by reconstructing the molecules and the side information simultaneously. In \textit{Experiment 1}, model selection was performed by continuously increasing the number of latent dimensions. Based on the knee of the mean absolute error (MAE), the appropriate number of latent archetypes was selected. 
In \textit{Experiments 2 and 3}, the number of latent dimensions was fixed to 19, corresponding to the optimal number of 20 archetypes from the model selection procedure. During training, the Lagrange multiplier $\lambda$ was steadily increased by increments of 1.01 every 500 iterations. For training, the Adam optimizer \citep{KingmaB14} was used, with an initial learning rate of 0.01. A learning rate decay was introduced, with an exponential decay of 0.95 every 10k iterations. The batch size was 2048 and the model was trained for a total of 350k iterations. The data set is divided in training and test set with a 90\%/10\% split. For visualization, the 3-dimensional molecular representations haven been created with \cite{jmol}.

\paragraph{Experiment 1: Model Selection.} MAE error is assessed while varying the number of archetypes. The result is shown in Figure \ref{fig:qm9selection}. Model selection is performed by observing for which number of archetypes the MAE starts to converge. The knee of this curve is used to select the optimal number of archetypes, which is 20. Obviously, if the number of archetypes is smaller, it becomes more difficult to reconstruct the data. This is explained by the fact that there exists a large number of molecules with \textit{very similar} heat capacities but at the same time \textit{distinctly different} geometric configurations. As a consequence, molecules with different configurations are mapped to archetypes with the similar heat capacity, making it hard to resolve the many--to--one mapping in the latent space. 

\paragraph{Experiment 2: Archetypal Molecules.} Archetypal mole\-cules are identified along with the heat capacities associated with them. A fixed number of 20 archetypes is used for optimal exploration-exploitation trade-off, in accordance with the model selection discussed in the previous section. In chemistry, the heat capacity at constant volume is defined as $C_v = \dfrac{d\epsilon}{dT} \bigm|_{v=const}$ where $\epsilon$ denotes the energy of a molecule and $T$ its temperature. This energy can be further decomposed into different parts, such that $\epsilon = \epsilon^{Tr}+\epsilon^R+\epsilon^V+\epsilon^E$. Each part is associated with a different degree of freedom of the system. Here, $Tr$ stands for translational, $R$ for rotational, $V$ for vibrational and $E$ for the electronic contributions to the total energy of the system \citep{atkins2010atkins, tinoco2002physical}. With this decomposition in mind, the different archetypal molecules associated with a particular heat capacity are compared in Figure \ref{fig:comparisions}. In both panels of that figure, the rows correspond to the three molecules in the QM9 data set (test set) that have been mapped closest to a vertex of the latent simplex and have thus been identified as being extremes with respect to the heat capacity. Out of a total of 20 vertices, molecules in close proximity to four of them are displayed here. Panel \ref{figure:case1} shows the configuration of six archetypal molecules. The upper three are all associated with a low heat capacity while the lower three all have a high heat capacity. This result can easily be interpreted, as the lower heat capacity can be traced back to the shorter chain length and the higher number of double bonds of these molecules, which makes them more stable and results in a lower vibrational energy $V$ and subsequently in a lower heat capacity. The inverse is observed for the linear archetypal molecules with higher heat capacities, which show, relative to their size, a lower number of double bonds and a long linear structure. Panel \ref{figure:case2} shows both linear (lower row) and non-linear archetypal molecules (upper row) but with similar atomic mass. Here, the non-linear molecules containing a cyclic structure in their geometry, are more stable and therefore have an overall slightly lower heat capacity compared to their linear counterparts of the same weight, shown in the second row.

\paragraph{Experiment 3: Interpolation Between Two Archetypal Molecules.} Interpolation is performed by plotting the samples from the test set which are closest to the connecting line between the two archetypes. As a result, one can observe a smooth transition from a molecule with a ring structure to a linear chain molecule. Both the starting and the end point of this interpolation is characterized by a similar heat capacity, such that these archetypes differ only in their geometric configuration but not with respect to their side information. As a consequence, any molecule in close proximity to that connecting line can differ only with respect to its structure, but \textit{must} display a similarly high heat capacity. Figure \ref{fig:interpolation} shows an example of such an interpolation. 

\begin{figure*}[!h]
    \centering
    \begin{subfigure}{0.45\textwidth}
        \centering
        \includegraphics[width=\textwidth]{long_chain_3d.png}
        \caption{}\label{figure:heatcap}
    \end{subfigure} \hspace{.35cm}
    \begin{subfigure}{0.45\textwidth}
        \centering
        \includegraphics[width=0.9\textwidth]{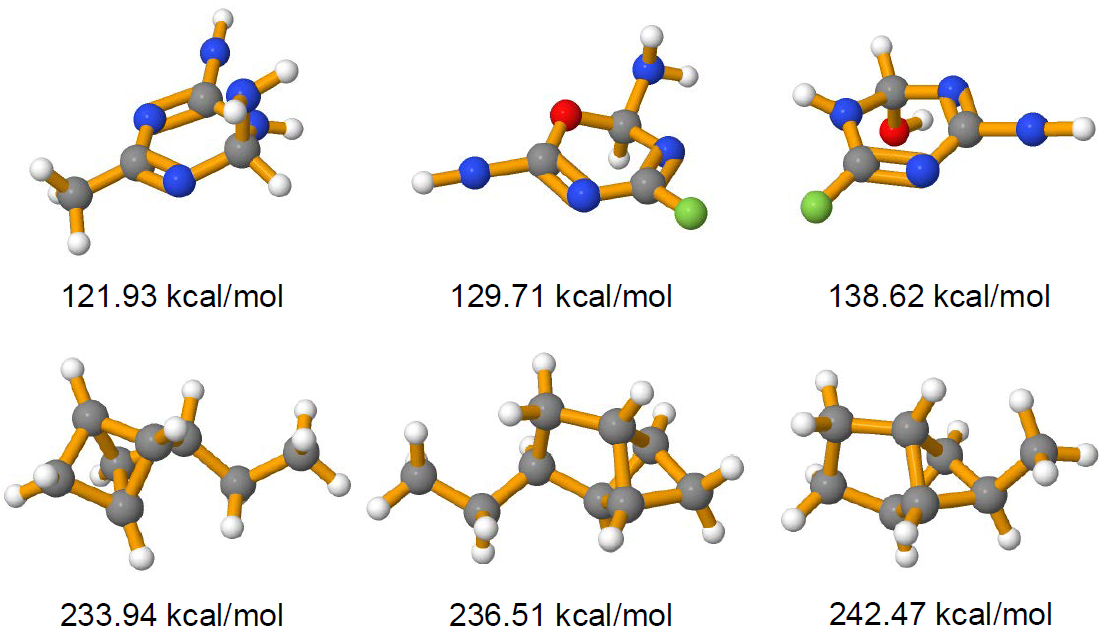}
        \caption{}\label{figure:gap}
    \end{subfigure}
    \caption{Panels (a) and (b) compare archetypal molecules identified using different side information: Here, the labels correspond to the heat capacity (panel a) and the band gap energy (panel b). The columns contain the three molecules of the test set closest to the given archetype.}    
    \label{fig:side}
\end{figure*}

\paragraph{Experiment 4: The Role Of Side Information And The Exploration Of Chemical Space.} Deep AA structures latent spaces both according to the information contained in the input to the encoder as well as the side information provided. As a consequence, any molecule characterized as a \textit{true} mixture of two or more arche\-types, given a specific side information such as \textit{heat capacity}, might suddenly be identified as archetypal should the side information change accordingly. In the following, archetypal molecules with respect to \textit{heat capacity} as the side information are compared to archetypes obtained while providing the \textit{band gap energy} of each molecule as the side information. In Figure \ref{figure:heatcap} archetypal molecules with both the highest and the lowest heat capacities are displayed while \ref{figure:gap} shows archetypes with highest and lowest band gap energies. The arche\-types significantly differ in their structure as well as their atomic composition. For example, arche\-typal mo\-le\-cules with low heat capacity are rather small, with only few C and O atoms, while archetypal molecules with a low band gap energy are characterized by ring structures containing N and H atoms. This illustrates how essential side information is for defining typicality but also for the subsequent interpretation of the obtained structure of the latent space.

\subsection{Alternative Priors For Deep Archetypal Analysis}
The standard normal distribution is a typical choice for the prior distribution $p(\mathbf{t})$ due to its simplicity and closed form solutions for the KL divergence.
However, employing alternative priors might be beneficial for the structure of the latent space and have an impact on the identified archetypes.

Leaving the wide range of well explored priors for vanilla VAEs aside, 
we explore a hierarchical prior that directly corresponds to the generative model of linear AA (Eq. \ref{eq:probAT_1}),
i.e. isotropic Gaussian noise around a linear combination of the archetypes:
\begin{align}\label{eq:prior_dir}
\mathbf{m} &\sim \text{Dir}_k(\boldsymbol{\alpha}=\mathbf{1}) \quad \text{ }\wedge\text{ } \quad
\mathbf{t} &\sim \mathcal{N}(\mathbf{m}Z^{\text{fixed}}, \mathbf{I})
\end{align}
We rely on Monte-Carlo sampling for the estimation of the KL divergence in Eq. \ref{eq:encoder_parametric}.
For comparing the different priors qualitatively, we run Deep AA on the Japanese Face Expressions with four archetypes.
The architecture used is similar to the previous experiments but we additionally learn the variance of the decoder.
The Lagrange parameters or weights in Eq. \ref{eq:ib_DeepAA} are set to 1000 for the archetype loss and to 100 for the KL divergence.

\begin{figure}[!ht]
\centering
\begin{subfigure}{0.48\textwidth}
   \includegraphics[width=1\linewidth]{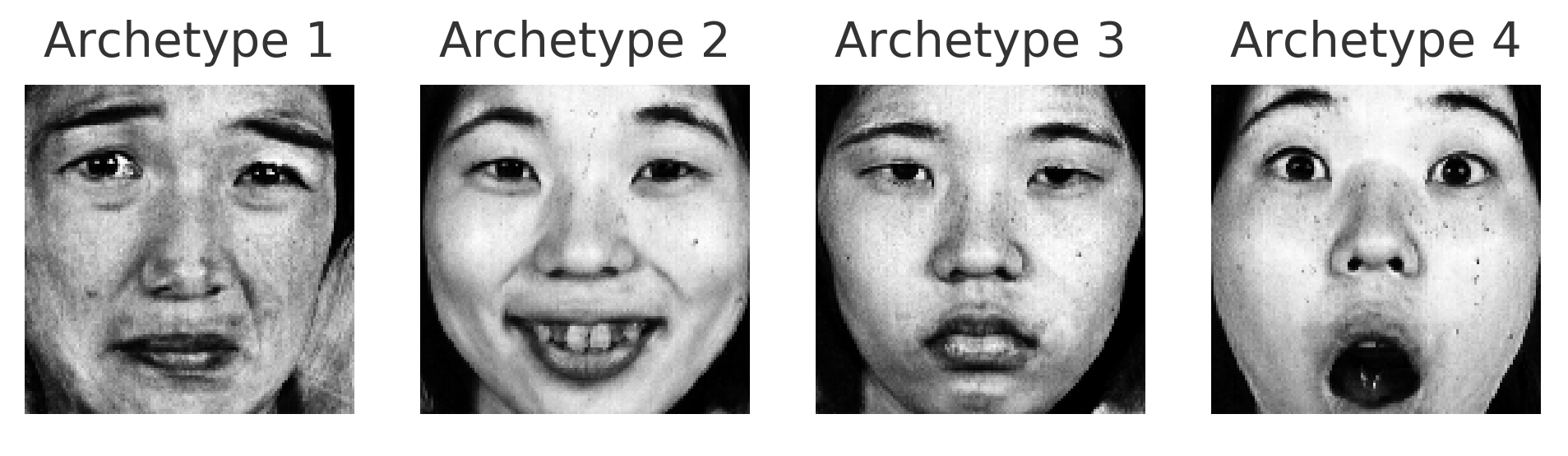}
   \caption{Archetypes learned using the sampling Dirichlet prior.}
\end{subfigure}\vspace{0.025\textwidth} 
\begin{subfigure}{0.48\textwidth}
   \includegraphics[width=1\linewidth]{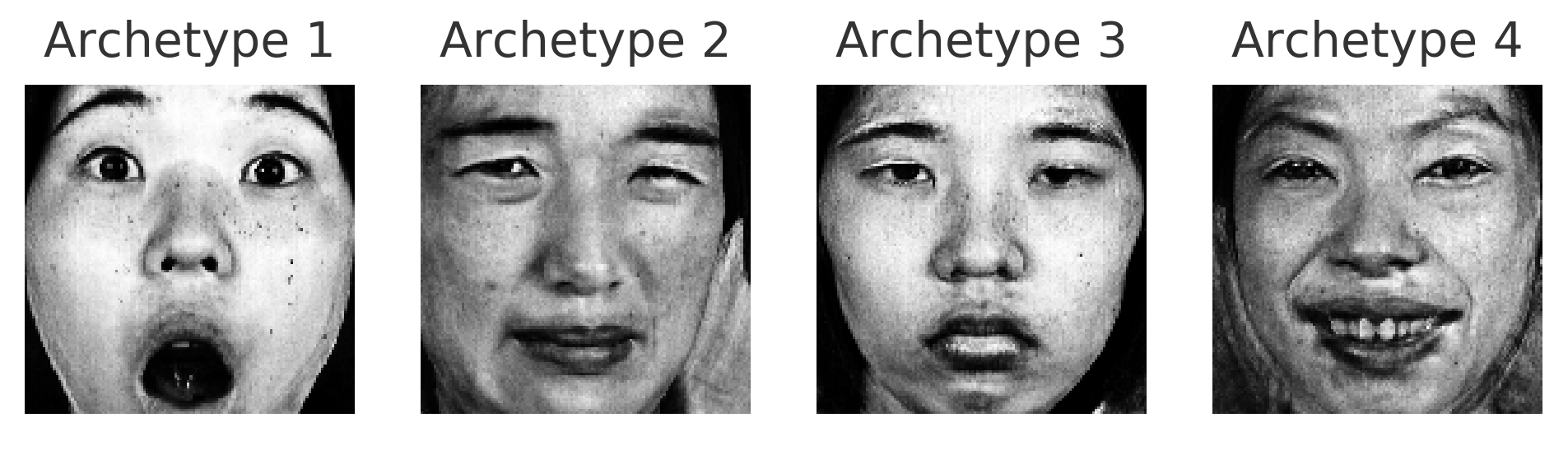}
   \caption{Archetypes learned using the standard normal prior.}
\end{subfigure}
\caption{Deep AA with $k$ = 4 archetypes using two different priors, which both identify similar archetypes.}
\label{fig:JAFFE:alt_prior}
\end{figure}

\begin{figure}[!ht]
\centering
\begin{subfigure}{0.44\textwidth}
   \includegraphics[width=1\linewidth]{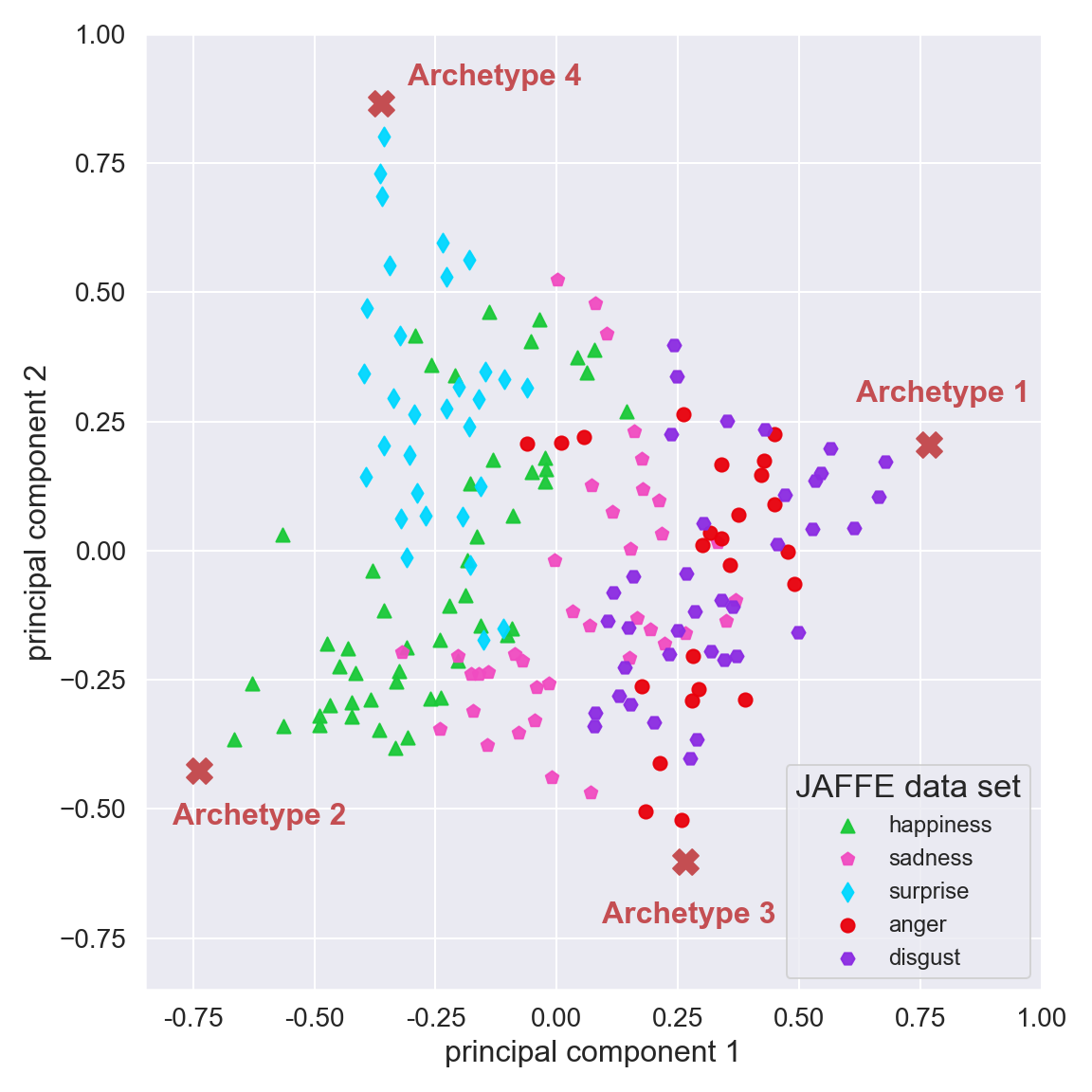}
   \caption{JAFFE latent space with sampling Dirichlet prior.}
   \label{fig:newDirPrior_latentSpace} 
\end{subfigure}\vspace{0.025\textwidth} 
\begin{subfigure}{0.44\textwidth}
   \includegraphics[width=1\linewidth]{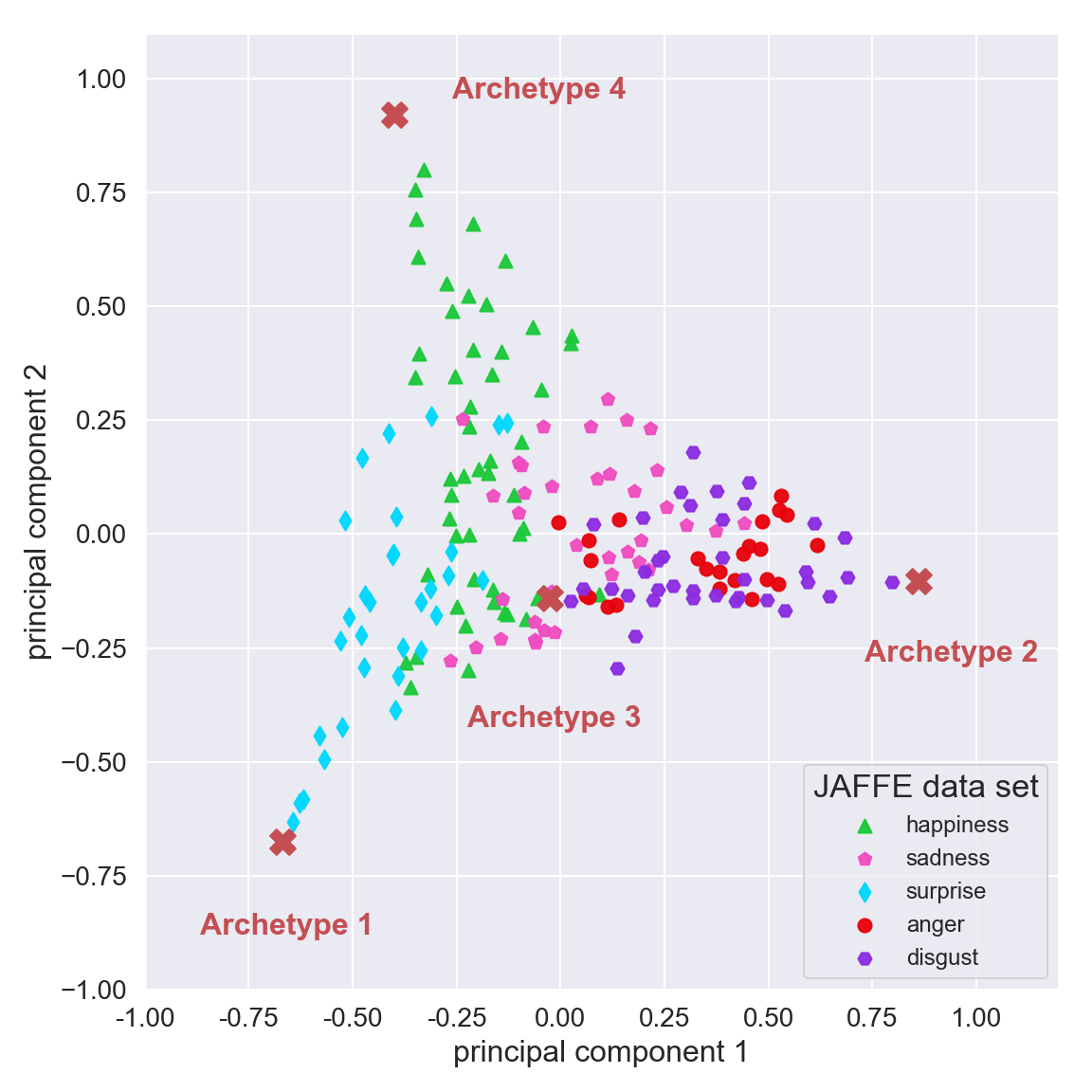}
   \caption{JAFFE latent space with standard normal prior.}
   \label{fig:oldGaussPrior_latentSpace}
\end{subfigure}
\caption{Latent spaces for the two different priors projected to their first two principal components using (linear) PCA. Explained variances: (a) $0.757$ (b) $0.74$.}
\label{fig:jaffe_interpolation}
\end{figure}

Figure \ref{fig:JAFFE:alt_prior} shows examples of the found archetypes for the standard normal prior and the sampling Dirichlet prior.
In general, different priors do not seem to strongly affect the found archetypes. However, the latent spaces do differ, which can be seen 
in the projection to the first two principal components in Figure \ref{fig:jaffe_interpolation}.
As a reference, a uniformly filled simplex would result in a triangle in the projection.
The difference is caused by large gaps in the higher-dimensional simplex when using the hierarchical prior,
which we assume is mainly due to the high variance estimation of the KL divergence.

In our experience, the choice of the prior is not of primary concern for finding archetypes,
as long as it encourages the latent space to be spread out inside the simplex -- be that via a standard normal, a uniform or the presented hierarchical prior.

%
%
%
%

%
%
\section{Conclusion}
\label{sec:5}

We have presented in this paper an extension of linear Archetypal Analysis, a technique for exploratory data analysis and interpretable machine learning. By performing Archetypal Analysis in the latent space of a Variational Autoencoder, we have demonstrated that the learned representation can be structured in a way that allows it to be characterized by its most extremal or archetypal representatives. As a result, each observation in a data set can be described as a convex mixture of these extremes. Endowed with such a structure, a latent space can be explored by varying the mixture coefficients with respect to the archetypes, instead of exploring the space by uniform sampling. Furthermore, we have demonstrated the need for including side information into the process of learning latent archetypal representations. As extremity is a form of typicality, a definition of what is understood to be \enquote{typical} needs to be given first. Providing such a definition is the role of side information and allows learning interpretable archetypes. In contrast to the original archetype model, our method offers three advantages: First, our model learns representations in a data-driven fashion, thereby reducing the need for expert knowledge. Second, our model can learn appropriate transformations to obtain meaningful archetypes, even if non-linear relations between features exist. Third, the incorporation of side information. 
The application of this new method is demonstrated on a sentiment analysis task, where emotion archetypes are identified based on female facial expressions, for which multi-rater based emotion scores are available as side information. A second application illustrates the exploration of the chemical space of small organic molecules and demonstrated how crucial side information is for interpreting the geometric configuration of these molecules.

\FloatBarrier

\bibliographystyle{spbasic}      
\bibliography{main.bib}   

\begin{thebibliography}{49}
\providecommand{\natexlab}[1]{#1}
\providecommand{\url}[1]{{#1}}
\providecommand{\urlprefix}{URL }
\expandafter\ifx\csname urlstyle\endcsname\relax
  \providecommand{\doi}[1]{DOI~\discretionary{}{}{}#1}\else
  \providecommand{\doi}{DOI~\discretionary{}{}{}\begingroup
  \urlstyle{rm}\Url}\fi
\providecommand{\eprint}[2][]{\url{#2}}

\bibitem[{Alemi et~al.(2016)Alemi, Fischer, Dillon, and Murphy}]{alemi2016}
Alemi AA, Fischer I, Dillon JV, Murphy K (2016) Deep variational information
  bottleneck. CoRR abs/1612.00410,
  \urlprefix\url{http://arxiv.org/abs/1612.00410}, \eprint{1612.00410}

\bibitem[{Anderson(1935)}]{anderson1935}
Anderson E (1935) The irises of the gaspe peninsula. Bulletin of the American
  Iris Society 59:2--5

\bibitem[{Atkins and de~Paula(2010)}]{atkins2010atkins}
Atkins P, de~Paula J (2010) Atkins' Physical Chemistry. OUP Oxford

\bibitem[{Bauckhage and Manshaei(2014)}]{bauck2014Kernel}
Bauckhage C, Manshaei K (2014) Kernel archetypal analysis for clustering web
  search frequency time series. In: 2014 22nd International Conference on
  Pattern Recognition, pp 1544--1549, \doi{10.1109/ICPR.2014.274}

\bibitem[{Bauckhage and Thurau(2009)}]{bauck2009ImgCol}
Bauckhage C, Thurau C (2009) Making archetypal analysis practical. In: Denzler
  J, Notni G, S{\"u}{\ss}e H (eds) Pattern Recognition, Springer Berlin
  Heidelberg, pp 272--281

\bibitem[{Bauckhage et~al.(2015)Bauckhage, Kersting, Hoppe, and
  Thurau}]{bauck2015FW}
Bauckhage C, Kersting K, Hoppe F, Thurau C (2015) Archetypal analysis as an
  autoencoder. In: Workshop New Challenges in Neural Computation 2015, pp
  8--16,
  \urlprefix\url{https://www.techfak.uni-bielefeld.de/~fschleif/mlr/mlr\_03\_2015.pdf}

\bibitem[{Blum and Reymond(2009)}]{gdb13}
Blum LC, Reymond JL (2009) 970 million druglike small molecules for virtual
  screening in the chemical universe database gdb-13. Journal of the American
  Chemical Society 131(25):8732--8733, \doi{10.1021/ja902302h}, pMID: 19505099

\bibitem[{Cabeza et~al.(2015)Cabeza, Gutierrez, Barreneche, Ushak, Fernandez,
  Fernadez, and Grageda}]{CABEZA20151106}
Cabeza LF, Gutierrez A, Barreneche C, Ushak S, Fernandez AG, Fernadez AI,
  Grageda M (2015) Lithium in thermal energy storage: A state-of-the-art
  review. Renewable and Sustainable Energy Reviews 42:1106 -- 1112

\bibitem[{Canhasi and Kononenko(2015)}]{can2015}
Canhasi E, Kononenko I (2015) Weighted hierarchical archetypal analysis for
  multi-document summarization. Computer Speech \& Language 37,
  \doi{10.1016/j.csl.2015.11.004}

\bibitem[{Cutler and Breiman(1994)}]{cutlerBreiman1994}
Cutler A, Breiman L (1994) Archetypal analysis. Technometrics 36(4):338--347,
  \doi{10.1080/00401706.1994.10485840},
  \urlprefix\url{http://digitalassets.lib.berkeley.edu/sdtr/ucb/text/379.pdf}

\bibitem[{Cutler and Stone(1997)}]{movingAT}
Cutler A, Stone E (1997) Moving archetypes. Physica D: Nonlinear Phenomena
  107(1):1--16, \doi{10.1016/s0167-2789(97)84209-1},
  \urlprefix\url{https://doi.org/10.1016/s0167-2789(97)84209-1}

\bibitem[{van Dijk et~al.(2019)van Dijk, Burkhardt, Amodio, Tong, Wolf, and
  Krishnaswamy}]{AAnet}
van Dijk D, Burkhardt D, Amodio M, Tong A, Wolf G, Krishnaswamy S (2019)
  Finding archetypal spaces for data using neural networks. arXiv preprint
  arXiv:190109078

\bibitem[{Djawdan et~al.(1996)Djawdan, Sugiyama, Schlaeger, Bradley, and
  Rose}]{drosophila}
Djawdan M, Sugiyama TT, Schlaeger LK, Bradley TJ, Rose MR (1996) Metabolic
  aspects of the trade-off between fecundity and longevity in drosophila
  melanogaster. Physiological Zoology 69(5):1176--1195

\bibitem[{El~Samad et~al.(2005)El~Samad, Khammash, Homescu, and
  Petzold}]{elSamad2005}
El~Samad H, Khammash M, Homescu C, Petzold L (2005) Optimal performance of the
  heat-shock gene regulatory network. Proceedings 16th IFAC World Congress 16,
  \urlprefix\url{https://engineering.ucsb.edu/~cse/Files/IFACC\_HS\_OPT04.pdf}

\bibitem[{Fisher(1936)}]{fisher1936}
Fisher RA (1936) The use of multiple measurements in taxonomic problems. Annals
  of Eugenics 7(Part II):179--188

\bibitem[{Garland(2014)}]{garland2014}
Garland TJJ (2014) Quick guides: Trade-offs. Current Biology 24(2):R60--R61

\bibitem[{Gomez-Bombarelli et~al.(2018)Gomez-Bombarelli, Wei, Duvenaud,
  Hernández-Lobato, Sánchez-Lengeling, Sheberla, Aguilera-Iparraguirre,
  Hirzel, Adams, and Aspuru-Guzik}]{Bombarelli}
Gomez-Bombarelli R, Wei JN, Duvenaud D, Hernández-Lobato JM,
  Sánchez-Lengeling B, Sheberla D, Aguilera-Iparraguirre J, Hirzel TD, Adams
  RP, Aspuru-Guzik A (2018) Automatic chemical design using a data-driven
  continuous representation of molecules. ACS Central Science 4(2):268--276

\bibitem[{H.~P.~Chan et~al.(2003)H.~P.~Chan, Mitchell, and Cram}]{chan2003}
H~P~Chan B, Mitchell D, Cram L (2003) Archetypal analysis of galaxy spectra.
  Monthly Notices of the Royal Astronomical Society 338,
  \doi{10.1046/j.1365-8711.2003.06099.x}

\bibitem[{Huggins et~al.(2007)Huggins, Pachter, and Sturmfels}]{huggins2007}
Huggins P, Pachter L, Sturmfels B (2007) Toward the human genotope. Bulletin of
  Mathematical Biology 69(8):2723--2735, \doi{10.1007/s11538-007-9244-7},
  \urlprefix\url{https://doi.org/10.1007/s11538-007-9244-7}

\bibitem[{{Jang} et~al.(2017){Jang}, {Gu}, and {Poole}}]{Jang}
{Jang} E, {Gu} S, {Poole} B (2017) {Categorical Reparameterization with
  Gumbel-Softmax}. International Conference on Learning Representations (ICLR)

\bibitem[{Jmol(2019)}]{jmol}
Jmol (2019) Jmol: an open-source java viewer for chemical structures in 3d
  \urlprefix\url{http://www.jmol.org/}

\bibitem[{Kaufmann et~al.(2015)Kaufmann, Keller, and Roth}]{kauf2015}
Kaufmann D, Keller S, Roth V (2015) Copula archetypal analysis. In: Gall J,
  Gehler P, Leibe B (eds) Pattern Recognition, Springer International
  Publishing, pp 117--128

\bibitem[{Kingma and Ba(2014)}]{KingmaB14}
Kingma DP, Ba J (2014) Adam: A method for stochastic optimization.
  abs/1412.6980

\bibitem[{Kingma and Welling(2013)}]{kingmaWelling2013}
Kingma DP, Welling M (2013) Auto-encoding variational bayes. CoRR abs/1312.6114

\bibitem[{Kingma et~al.(2014)Kingma, Mohamed, Rezende, and
  Welling}]{KingmaSemi}
Kingma DP, Mohamed S, Rezende DJ, Welling M (2014) Semi-supervised learning
  with deep generative models. In: Advances in Neural Information Processing
  Systems 27: Annual Conference on Neural Information Processing Systems 2014,
  December 8-13 2014, Montreal, Quebec, Canada, pp 3581--3589

\bibitem[{Kirkpatrick and Ellis(2004)}]{kirkpatrick2004}
Kirkpatrick P, Ellis C (2004) Chemical space. Nature 432(823):1476--4687,
  \doi{10.1038/432823a}

\bibitem[{Lyons et~al.(1998)Lyons, Akamatsu, Kamachi, and Gyoba}]{lyons1998}
Lyons MJ, Akamatsu S, Kamachi M, Gyoba J (1998) Coding facial expressions with
  gabor wavelets. 3rd IEEE International Conference on Automatic Face and
  Gesture Recognition pp 200--205, \doi{10.1109/AFGR.1998.670949},
  \urlprefix\url{https://zenodo.org/record/3430156}

\bibitem[{Mair and Brefeld(2019)}]{at-coresets}
Mair S, Brefeld U (2019) Coresets for archetypal analysis. In: Advances in
  Neural Information Processing Systems 32, Curran Associates, Inc., pp
  7245--7253,
  \urlprefix\url{http://papers.nips.cc/paper/8945-coresets-for-archetypal-analysis.pdf}

\bibitem[{Miller(1953)}]{miller53}
Miller SL (1953) A production of amino acids under possible primitive earth
  conditions. Science pp 528--529

\bibitem[{M{\o}rup and Hansen(2012)}]{morupKernelAA}
M{\o}rup M, Hansen LK (2012) Archetypal analysis for machine learning and data
  mining. Neurocomputing 80:54--63

\bibitem[{Norberg et~al.(1987)Norberg, Rayner, and Lighthill}]{norberg1987}
Norberg UM, Rayner JMV, Lighthill MJ (1987) Ecological morphology and flight in
  bats (mammalia; chiroptera): wing adaptations, flight performance, foraging
  strategy and echolocation. Philosophical Transactions of the Royal Society of
  London B, Biological Sciences 316(1179),
  \urlprefix\url{https://royalsocietypublishing.org/doi/abs/10.1098/rstb.1987.0030}

\bibitem[{{Parbhoo} et~al.(2018){Parbhoo}, {Wieser}, and
  {Roth}}]{2018arXiv180702326P}
{Parbhoo} S, {Wieser} M, {Roth} V (2018) {Causal Deep Information Bottleneck}.
  arXiv e-prints arXiv:1807.02326, \eprint{1807.02326}

\bibitem[{Prabhakaran et~al.(2012)Prabhakaran, Raman, Vogt, and
  Roth}]{sand2012}
Prabhakaran S, Raman S, Vogt JE, Roth V (2012) Automatic model selection in
  archetype analysis. In: Pinz A, Pock T, Bischof H, Leberl F (eds) Pattern
  Recognition, Springer Berlin Heidelberg, pp 458--467

\bibitem[{Ramakrishnan et~al.(2014)Ramakrishnan, Dral, Rupp, and von
  Lilienfeld}]{rama2014}
Ramakrishnan R, Dral PO, Rupp M, von Lilienfeld OA (2014) Quantum chemistry
  structures and properties of 134 kilo molecules. Scientific Data 1

\bibitem[{Rezende and Mohamed(2015)}]{pmlrrezende15}
Rezende D, Mohamed S (2015) Variational inference with normalizing flows. In:
  Bach F, Blei D (eds) Proceedings of the 32nd International Conference on
  Machine Learning, PMLR, Lille, France, Proceedings of Machine Learning
  Research, vol~37, pp 1530--1538

\bibitem[{Rezende et~al.(2014)Rezende, Mohamed, and Wierstra}]{rezende2014}
Rezende DJ, Mohamed S, Wierstra D (2014) Stochastic backpropagation and
  approximate inference in deep generative models 32(2):1278--1286

\bibitem[{Ruddigkeit et~al.(2012)Ruddigkeit, van Deursen, Blum, and
  Reymond}]{rudd2012}
Ruddigkeit L, van Deursen R, Blum LC, Reymond JL (2012) Enumeration of 166
  billion organic small molecules in the chemical universe database gdb-17.
  Journal of Chemical Information and Modeling 52(11):2864--2875,
  \doi{10.1021/ci300415d},
  \urlprefix\url{https://pubs.acs.org/doi/10.1021/ci300415d}, pMID: 23088335

\bibitem[{Schuetz et~al.(2012)Schuetz, Zamboni, Zampieri, Heinemann, and
  Sauer}]{schuetz2012}
Schuetz R, Zamboni N, Zampieri M, Heinemann M, Sauer U (2012) Multidimensional
  optimality of microbial metabolism. Science (New York, NY) 336:601--4,
  \doi{10.1126/science.1216882}

\bibitem[{Seth and Eugster(2016)}]{seth2016}
Seth S, Eugster MJA (2016) Probabilistic archetypal analysis. Machine Learning
  102(1):85--113, \doi{10.1007/s10994-015-5498-8},
  \urlprefix\url{https://doi.org/10.1007/s10994-015-5498-8}

\bibitem[{Shoval et~al.(2012)Shoval, Sheftel, Shinar, Hart, Ramote, Mayo,
  Dekel, Kavanagh, and Alon}]{shoval2012}
Shoval O, Sheftel H, Shinar G, Hart Y, Ramote O, Mayo A, Dekel E, Kavanagh K,
  Alon U (2012) Evolutionary trade-offs, pareto optimality, and the geometry of
  phenotype space. Science 336(6085):1157--1160, \doi{10.1126/science.1217405},
  \urlprefix\url{http://science.sciencemag.org/content/336/6085/1157},
  \eprint{http://science.sciencemag.org/content/336/6085/1157.full.pdf}

\bibitem[{Steinbeck et~al.(2003)Steinbeck, Han, Kuhn, Horlacher, Luttmann, and
  Willighagen}]{cdk}
Steinbeck C, Han YQ, Kuhn S, Horlacher O, Luttmann E, Willighagen E (2003) {The
  Chemistry Development Kit (CDK): An open-source Java library for chemo- and
  bioinformatics}. {Journal of Chemical Information and Computer Sciences}
  43(2):493--500

\bibitem[{Steuer(1986)}]{steuer1986}
Steuer R (1986) Multiple Criteria Optimization: Theory, Computation and
  Application. John Wiley \& Sons

\bibitem[{Stone and Cutler(1996)}]{spatiotempAT}
Stone E, Cutler A (1996) Introduction to archetypal analysis of spatio-temporal
  dynamics. Physica D: Nonlinear Phenomena 96(1-4):110--131,
  \doi{10.1016/0167-2789(96)00016-4},
  \urlprefix\url{https://doi.org/10.1016/0167-2789(96)00016-4}

\bibitem[{Tendler et~al.(2015)Tendler, Mayo, and Alon}]{tendler2015}
Tendler A, Mayo A, Alon U (2015) Evolutionary tradeoffs, pareto optimality and
  the morphology of ammonite shells. BMC Systems Biology 9(1),
  \doi{10.1186/s12918-015-0149-z},
  \urlprefix\url{https://doi.org/10.1186/s12918-015-0149-z}

\bibitem[{Tinoco(2002)}]{tinoco2002physical}
Tinoco I (2002) Physical Chemistry: Principles and Applications in Biological
  Sciences. No. S. 229-313 in Physical Chemistry: Principles and Applications
  in Biological Sciences, Prentice Hall

\bibitem[{Tishby et~al.(2000)Tishby, Pereira, and Bialek}]{tishby2000}
Tishby N, Pereira FC, Bialek W (2000) The information bottleneck method. arXiv
  preprint physics/0004057

\bibitem[{Visini et~al.(2017)Visini, Arus-Pous, Awale, and
  Reymond}]{visini2017}
Visini R, Arus-Pous J, Awale M, Reymond JL (2017) Virtual exploration of the
  ring systems chemical universe. Journal of Chemical Information and Modeling
  57(11):2707--2718, \doi{10.1021/acs.jcim.7b00457},
  \urlprefix\url{https://doi.org/10.1021/acs.jcim.7b00457}, pMID: 29019686,
  \eprint{https://doi.org/10.1021/acs.jcim.7b00457}

\bibitem[{{Wieczorek} et~al.(2018){Wieczorek}, {Wieser}, {Murezzan}, and
  {Roth}}]{2018arXiv180406216W}
{Wieczorek} A, {Wieser} M, {Murezzan} D, {Roth} V (2018) {Learning Sparse
  Latent Representations with the Deep Copula Information Bottleneck}.
  International Conference on Learning Representations (ICLR)

\bibitem[{Wynen et~al.(2018)Wynen, Schmid, and Mairal}]{styleAA}
Wynen D, Schmid C, Mairal J (2018) Unsupervised learning of artistic styles
  with archetypal style analysis. In: Advances in Neural Information Processing
  Systems, pp 6584--6593

\end{thebibliography}


\end{document}